\documentclass{article}

\usepackage{makecell}
\usepackage[final]{lg}
\usepackage[utf8]{inputenc}
\usepackage[T1]{fontenc}
\usepackage[pdfencoding=auto, psdextra, unicode]{hyperref}
\usepackage{url}
\usepackage{booktabs}
\usepackage{rotating}
\usepackage{amsfonts}
\usepackage{amsmath}
\usepackage{nicefrac}
\usepackage[dvipsnames]{xcolor}
\usepackage{adjustbox}
\usepackage{multirow}
\usepackage{amssymb}
\usepackage{longtable}
\usepackage{comment}
\usepackage[english]{babel}
\usepackage[autostyle]{csquotes}
\usepackage{hhline}
\usepackage{tabu}
\usepackage{tablefootnote}
\usepackage{graphicx}
\usepackage{wrapfig}
\usepackage{array}
\usepackage{caption}
\usepackage{tcolorbox}
\usepackage{relsize}
\usepackage{colortbl}
\tcbuselibrary{breakable}
\usepackage{floatpag}
\floatpagestyle{plain}
\usepackage{float}
\usepackage{algorithm}
\usepackage{algpseudocode}
\usepackage{threeparttable}
\usepackage{enumitem}
\setlist[enumerate,itemize]{leftmargin=1.5em, topsep=1pt, partopsep=0pt,
                            parsep=0pt, itemsep=0.5pt}
\usepackage{placeins}

\usepackage{bm}
\usepackage{tikz}
\usetikzlibrary{positioning,arrows.meta,calc,fit,backgrounds}
\usepackage{nicematrix}

\definecolor{LightYellow}{rgb}{1.0,0.98,0.85}
\definecolor{LightPurple}{rgb}{0.957,0.918,0.987}
\definecolor{LightOrange}{rgb}{0.992,0.933,0.867}
\definecolor{Band}{gray}{0.90}

\newtcolorbox{HeroCard}{
  breakable,
  colback=black!3,
  colframe=black!3,
  boxrule=0pt,
  arc=3mm,
  left=12pt,right=12pt,top=7pt,bottom=7pt
}

\definecolor{MainPurple}{HTML}{A451E4}

\makeatletter
\edef\OrigRM{\rmdefault}
\edef\OrigSF{\sfdefault}
\edef\OrigTT{\ttdefault}
\renewcommand{\@maketitle}{%
  \begin{center}
  \begin{HeroCard}
  {%
    \fontfamily{put}\selectfont

    {\LARGE\bfseries \@title\par}
    \vspace{0.5em}

    {\normalsize \@author\par}
    \vspace{0.7em}

    {\normalsize
    \setlength{\baselineskip}{1.25\baselineskip}
    \noindent Time series foundation models (TSFMs) are pretrained on series from diverse domains, where
demand series make up only a small fraction. Demand data has properties that such corpora rarely
contain: Short histories, frequent zeros, censoring by stock-outs, and exogenous events that the
series does not record. To this end, we propose \textbf{EXAONE Demand}, built on 1) a
\textit{demand-specific corpus} and 2) a \textit{demand-aware adapter}. For the corpus, we assemble 11.3M series and 48.4B observations from 73 sources, and a
\textit{synthetic generator} supplies the behaviour that open demand data under-represents. For the
adapter, we attach low-rank branches to a frozen general-domain backbone, one for each of the four
demand classes (smooth, intermittent, erratic, and lumpy), and a router that reads eight
scale-free statistics of the input series decides how much each branch contributes. We build EXAONE Demand in two
versions, one trained on real-world and synthetic demand together and one trained on the
synthetic corpus alone. On 22 held-out datasets, both versions outperform 36 TSFMs,
and real-world demand adds a gain over synthetic data alone.

\vspace{0.55em}
\noindent
{\footnotesize
\textbf{Model}\quad\url{https://huggingface.co/LG-AI-Research/EXAONE-Demand-1.0}\\[1pt]
\textbf{Code}\quad\;\url{https://github.com/LGAI-Research/EXAONE-Forecast}
}
\par}

  }%
  \end{HeroCard}
  \end{center}

  \@thanks
  \global\let\@thanks\@empty
  \setcounter{footnote}{0}
  \vspace{0.5em}
}%
\makeatother

\usepackage{fourier}
\renewcommand{\rmdefault}{\OrigRM}
\renewcommand{\sfdefault}{\OrigSF}
\renewcommand{\ttdefault}{\OrigTT}

\usepackage{xspace}

\newcommand{\comp}{LG~AI~Research}

\newcommand{\best}[1]{\textcolor{red}{\textbf{#1}}}
\newcommand{\second}[1]{\textcolor{blue}{\underline{#1}}}

\title{
\includegraphics[height=5.5mm]{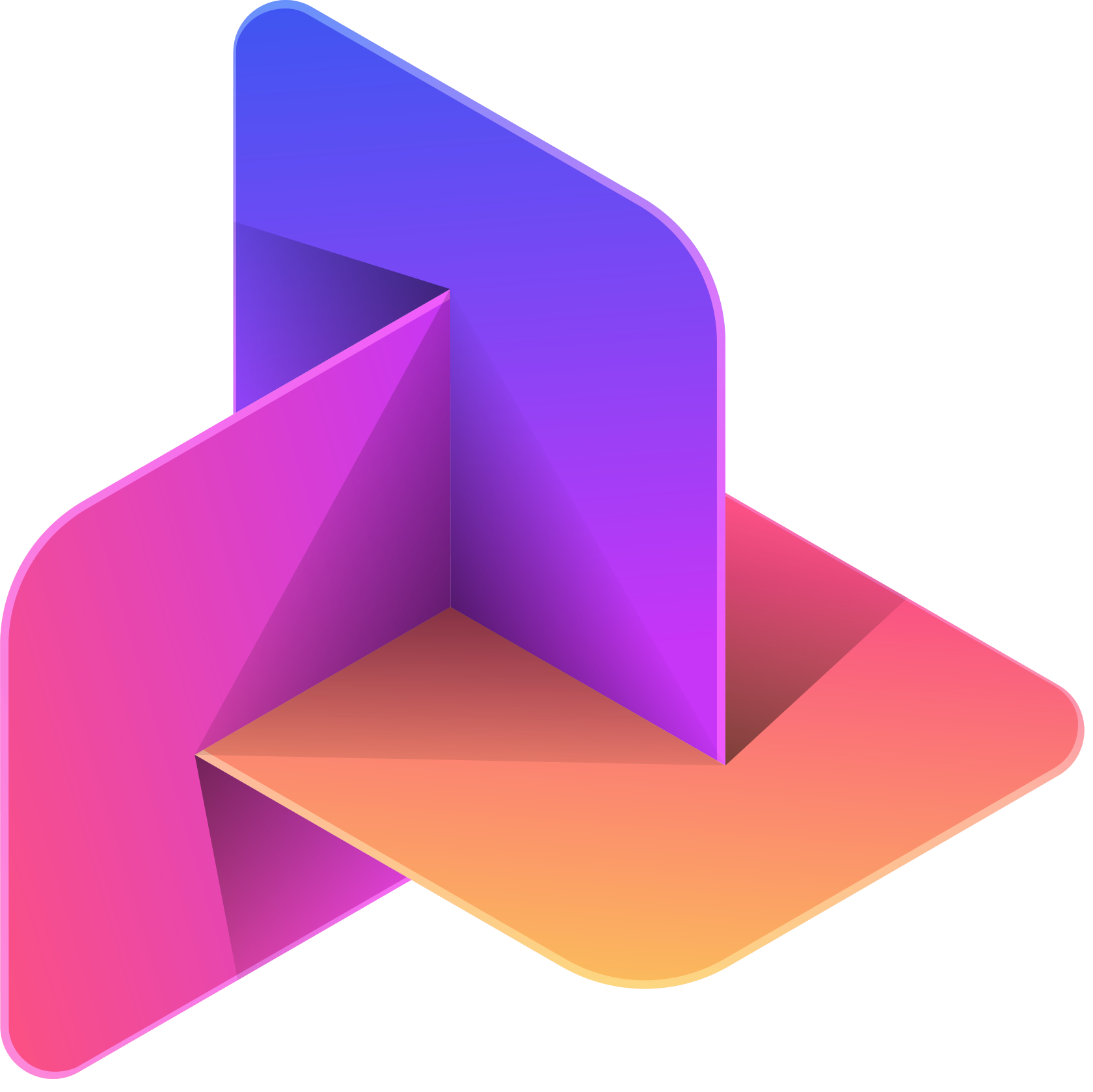}
EXAONE Demand 1.0:\\ 
\large{A Time Series Foundation Model for Demand Forecasting}
}

\author{%
  \comp$^{*}$ \\[2pt]
  {\normalsize Seunghan Lee, Sangjun Han, Jun Seo, Junhyeok Kang, Jaehoon Lee, Tae Yoon Lim, \\
  Dongwan Kang, Hwanil Choi, Minjae Kim, Sungdong Yoo, Soonyoung Lee, Wonbin Ahn$^{\dagger}$}
}

\makeatletter
\newcommand{\blfootnote}[1]{%
  \begingroup
    \renewcommand{\thefootnote}{}\footnote{#1}%
    \addtocounter{footnote}{-1}%
  \endgroup
}
\makeatother

\begin{document}

\maketitle
\blfootnote{$^{\dagger}$\,Corresponding author.}

\section{Introduction}
\label{sec:intro}

Demand forecasting predicts how much of a product, a service, or a resource will be requested over
a future period, and that prediction decides what an organisation stocks, staffs, and
ships~\citep{makridakis2022m5}. Its
two directions of error carry different costs, where \textit{under-forecasting} leaves demand unmet and
\textit{over-forecasting} commits resources that are never used. Demand
behaves unlike the \textit{smooth} and \textit{long} series that dominate general forecasting
benchmarks, and it is therefore studied as a problem of its own, with its own error measures and its
own taxonomy of series~\citep{syntetos2005categorization}.

Time series foundation models (TSFMs) are pretrained at scale to forecast across
domains~\citep{ansari2024chronos,woo2024unified,das2024decoder,goswami2024moment}, and they reach
strong zero-shot accuracy on general benchmarks~\citep{godahewa2021monash,aksu2024gifteval}.
Their corpora are assembled for \textit{breadth} rather than for \textit{demand},
and the properties that make demand hard are diluted by the far more numerous other series. We argue that \textit{demand forecasting needs a corpus of its own}, one that reflects
these properties instead of averaging them out.

\begin{figure}[t]
\centering
\begin{minipage}[t]{0.547\linewidth}
\centering
\vspace{0pt}
\includegraphics[width=\linewidth]{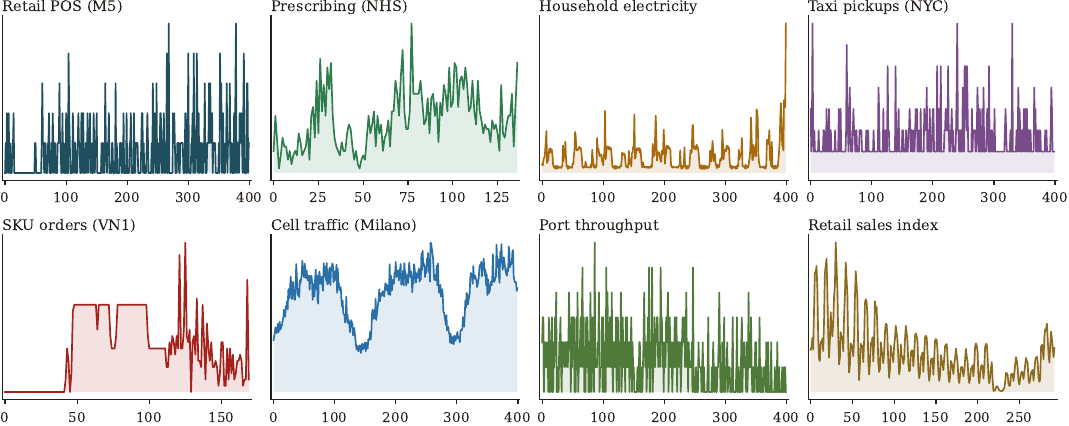}
\caption{\textbf{Representative real-world demand series.} One series is drawn per source from the
corpus. Retail and SKU-level series are dominated by zeros and irregular spikes, while
energy and transport series are long with strong daily and weekly cycles.}
\label{fig:real}
\end{minipage}\hfill
\begin{minipage}[t]{0.425\linewidth}
\centering
\vspace{0pt}
\includegraphics[width=0.97\linewidth]{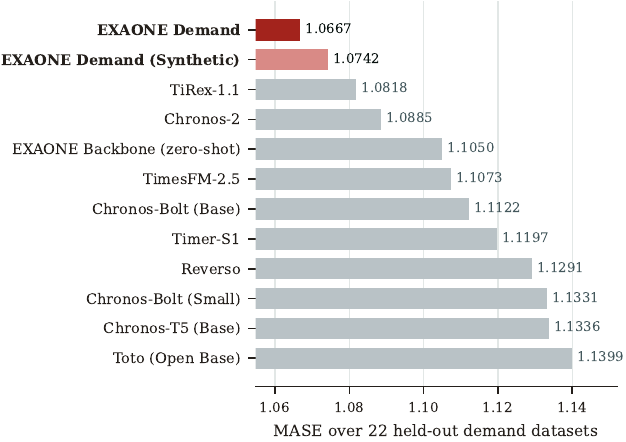}
\vspace{-1pt}
\caption{Performance on demand datasets.}
\label{fig:teaser}
\end{minipage}
\end{figure}

\textbf{Three properties of demand data.}
Three properties recur across our demand series and are rare in general corpora, and
Figure~\ref{fig:real} shows all three in eight series, one drawn from each of eight open sources
of the corpus.
\begin{itemize}[topsep=1pt,itemsep=0.5pt,parsep=0pt]
\item \textbf{Intermittency.} A large share of series are \textit{zero} most of the time, broken by
irregular non-zero periods, which the \textit{average demand interval} (ADI) and the \textit{squared
coefficient of variation} (CV$^2$) characterise~\citep{syntetos2005categorization}.
\item \textbf{Shortness.} Demand is recorded at the granularity at which it is \textit{ordered},
which is weekly or monthly per SKU and per store, leaving individual series far shorter than the
ones general benchmarks are built on.
\item \textbf{Exogenous and censored structure.} Promotions, holidays, and life-cycles drive level
shifts whose cause is \textit{absent from the series}, and observed sales are a \textit{censored}
view of latent demand whenever inventory runs out.
\end{itemize}

A demand corpus does not hold one kind of series but four, where \textit{smooth},
\textit{intermittent}, \textit{erratic}, and \textit{lumpy} demand arrive mixed together. They ask different questions of a forecaster, where \textit{intermittency} turns the question into
\textit{when} demand arrives, while \textit{erratic} and \textit{lumpy} behaviour turns it into
\textit{how much} arrives at once. A corpus that holds all four is
not enough by itself, since one set of weights fitted to the mixture has to serve both questions,
and moving it toward one behaviour costs accuracy on the other, a trade-off that a single adapter has no means to resolve.

\begin{wrapfigure}{r}{0.408\linewidth}
\centering
\vspace{-1pt}
\includegraphics[width=\linewidth]{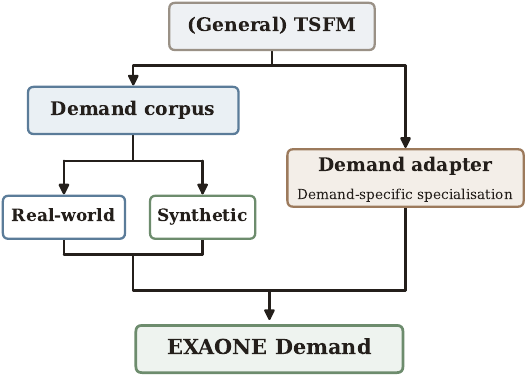}
\caption{Overview of EXAONE Demand.}
\label{fig:overview}
\vspace{-10pt}
\end{wrapfigure}

To this end, we propose \textbf{EXAONE Demand}, a TSFM for demand forecasting that
\textit{specialises by demand behaviour}. As Figure~\ref{fig:overview} lays out, it rests on two
components: 1) A \textit{demand-specific corpus} built from real-world demand sources and a synthetic
generator, and 2) a \textit{demand-aware adapter} that mixes low-rank branches by the behaviour of
the input series. Specifically, we adapt a general-domain TSFM pretrained on
KernelSynth~\citep{ansari2024chronos} alone, namely synthetic series drawn from Gaussian process
kernels,
keeping its weights frozen and training only five low-rank branches, one shared by every series and
four tied to the demand classes, where a \textit{router} decides how much each of the four
contributes. We build EXAONE Demand in two versions, one trained on
real-world and synthetic demand together and one trained on the synthetic corpus alone. As shown in Figure~\ref{fig:teaser}, both versions
outperform the state-of-the-art (SoTA) TSFMs on 22 held-out demand datasets, with real-world demand adding a gain over synthetic data.
%

\section{Construction of Demand Dataset}
\label{sec:data}

\begin{figure}[t]
\centering
\includegraphics[width=\linewidth]{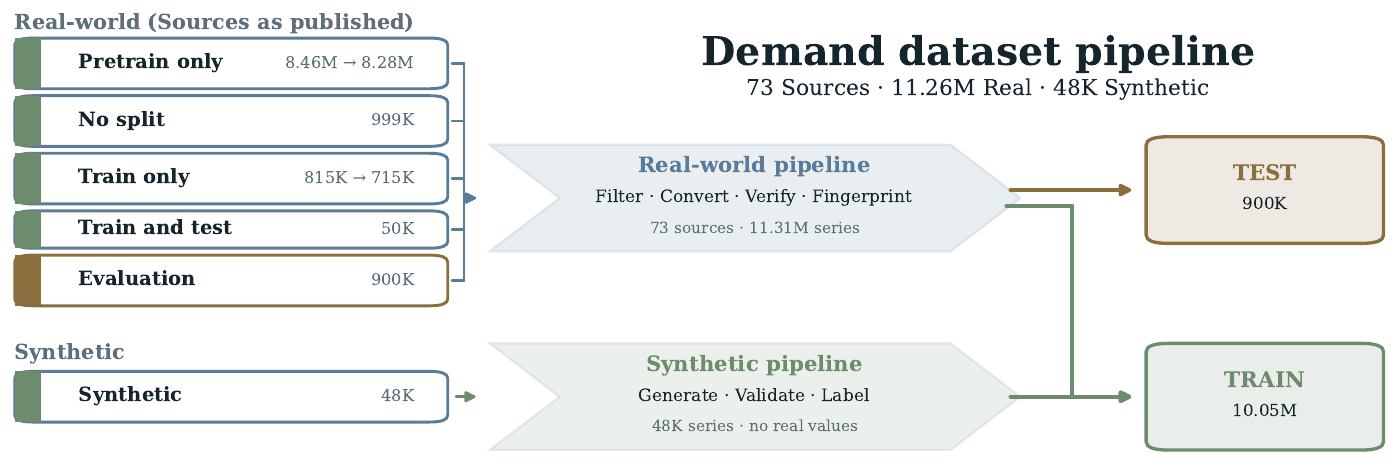}
\caption{\textbf{Pipeline for demand dataset construction.}
\textbf{1) Collection:} Real sources are grouped by their published role, and the synthetic corpus
forms a group of its own below.
\textbf{2) Preprocessing:} The two corpora run on separate tracks. The real track filters, converts,
and verifies each source before removing leaked series, while the synthetic track generates,
validates, and labels its own.
\textbf{3) Train-test split:} Of the five roles, only the evaluation suites reach the test side, and
the other four are used for training, together with every one of the synthetic series.}
\label{fig:pipeline}
\end{figure}

We construct the demand dataset in the following three steps, as illustrated in
Figure~\ref{fig:pipeline}:
\begin{itemize}[topsep=1pt,itemsep=0.5pt,parsep=0pt]
\item \textbf{Collection} (\S\ref{sec:collect}): We gather various sources and keep only the series that measure demand.
\item \textbf{Preprocessing} (\S\ref{sec:prep}): We convert every source into a common schema and index every series we keep.
\item \textbf{Train-test split} (\S\ref{sec:split}): We assign sources by their published role and remove the training series that leak.
\end{itemize}

A synthetic corpus is built in parallel, and the real and synthetic corpora stay apart until the last step,
where every synthetic series is used for training. Section~\ref{sec:synth} describes the design and the
generating process behind it.

\begin{table}[t]
\centering
\caption{\textbf{Collected sources by domain.} The count is over the 93 collected sources, and the
last two rows are source types rather than domains, since each of them holds many datasets under
a single published name.}
\label{tbl:sources}
\vspace{1mm}
\begin{NiceTabular}{l | r | X[l]}[width=\linewidth]
\toprule
\textbf{Domain} & \textbf{Sources} & \textbf{Representative content} \\
\midrule
Retail and e-commerce & 24 & Store and SKU sales, grocery baskets, fashion and online retail \\
Energy                & 23 & Household and industrial electricity load, national demand \\
Transport             & 14 & Taxi and ride-hailing pickups, bike share, transit ridership, port calls \\
Healthcare            &  6 & Prescription dispensing at practice, pharmacy, and hospital level \\
Service and hospitality & 4 & Restaurant covers, hotel and booking demand \\
Telecommunications    &  3 & Cell-grid traffic \\
Economics             &  3 & Retail sales indices, industrial production, trade volumes \\
Web and compute       &  2 & Page requests, virtual-machine resource usage \\
General-purpose corpora & 7 & Time-300B, GIFT-Eval, LOTSA, Chronos, fev-bench, Monash \\
Competition benchmarks  & 7 & M4, M5, VN1, Rossmann, Favorita, and related \\
\bottomrule
\end{NiceTabular}
\end{table}

\subsection{Collection}
\label{sec:collect}

The first decision is not \textit{where} to find data but \textit{what} to accept as demand. This
cannot be taken from the data, as nothing in a published file marks a series as demand and no
metadata field records it. We therefore state a definition of our own, taking a demand series to be
a measurement of \emph{consumption, sales, or utilisation per unit of time by an identifiable
entity}. What the definition rejects matters as much as what it keeps, since the rejected series are
often the ones that look most like demand. Three rejections came up in almost every source, and we state them as rules:

\begin{itemize}[topsep=1pt,itemsep=0.5pt,parsep=0pt]
\item \textbf{\textit{1) Supply is not demand.}} We keep a series only when it measures what was taken, not
what was produced. Solar and wind generation are therefore excluded, and so is the \texttt{export} reading of a residential smart meter, while the \texttt{import} reading of the same meter is kept, since it records what the household
actually drew.
\item \textbf{\textit{2) Physical quantities are not demand.}} Transformer oil temperature, air quality,
river flow, and weather measure the state of a system rather than anyone's use of it. Several of them are
standard forecasting benchmarks, and they therefore reach us inside the broad collections described below
rather than as sources of their own.
\item \textbf{\textit{3) Inventory is not demand.}} The number of bikes available at a docking station is
supply, and the number of trips starting there is demand. The same file often holds both, and we
take only the second.
\end{itemize}

Two cases are not decided by the rules above, since the question there is not whether a series is
demand but what a single number in it stands for.
(1) \textbf{Sources with no series:} Event logs such as taxi trips, bike rentals, and
municipal service requests store one row per event. They become demand
only after the events are counted per entity and per time step, which makes those two settings the
definition of demand for that source.
(2) \textbf{Sources with several candidate values:} In pharmaceutical data we take prescription
\emph{items} rather than \emph{quantity}, since quantity is measured in tablets for one drug and in millilitres for another, and the two cannot be summed on a common scale.

\textbf{Sources.}
Demand data is published in three types of sources, and relying on a single type yields a
biased corpus, as benchmark suites and public data archives provide almost disjoint material. We
therefore collect from all three:

\begin{itemize}[topsep=1pt,itemsep=0.5pt,parsep=0pt]
\item \textbf{Type 1) Forecasting benchmarks and competitions.} M4, M5, VN1, Rossmann, Favorita, and related retail
competitions, which define the conventional evaluation of demand forecasting in the literature.
\item \textbf{Type 2) General-purpose corpora.} Broad collections assembled by other groups to cover time series
in general, from which we keep only the subsets that are relevant to demand and discard the rest.
\item \textbf{Type 3) Domain-specific open data.} National health services, transport authorities, energy
regulators, and statistical agencies, which publish demand directly but are absent from forecasting
benchmarks.
\end{itemize}

\begin{wrapfigure}{r}{0.48\linewidth}
\vspace{-\intextsep}
\vspace{1pt}
\centering
\includegraphics[trim=19bp 19bp 4bp 19bp, clip, width=\linewidth]{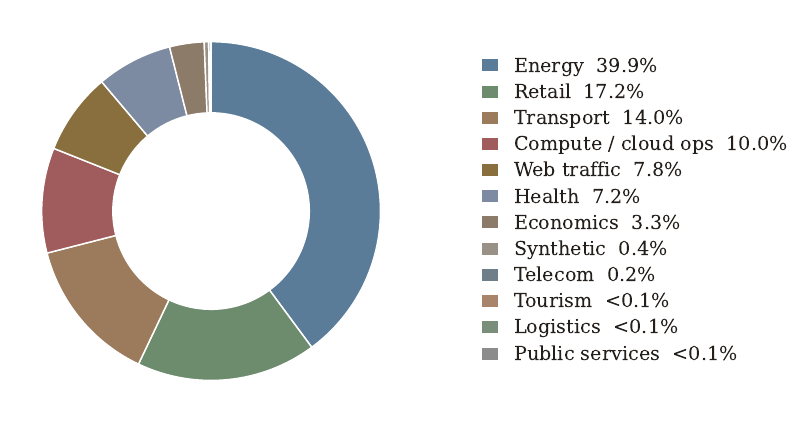}
\vspace{-17pt}
\caption{\textbf{Corpus composition by series count.} Every series carries one domain of the
converted corpus.}
\label{fig:domains}
\vspace{-12pt}
\end{wrapfigure}

We collect 93 sources in total, of which 73 are converted to a common schema. The remaining 20 are
excluded because 1) their contents are not tabular, 2) they duplicate sources already collected, or
3) they require credentials we do not hold. Table~\ref{tbl:sources} summarises the collection by
domain, and Figure~\ref{fig:domains} shows the distribution of the converted corpus.

\textbf{Published roles.}
\label{sec:roles}
Each source is released with a role determined by its publisher, which we record rather than assign.
We identify five roles: 1) \textit{Pretrain only}, corpora with no held-out part, 2) \textit{No
split}, releases published as a single undivided file, 3) \textit{Train only}, releases that publish
a training file while keeping the held-out part private, 4) \textit{Train and test}, releases that
ship a fixed train and test pair, and 5) \textit{Evaluation}, suites published to be scored against.
Table~\ref{tbl:roles} lists the sources in each role, and \S\ref{sec:split} describes how each role
is used in training and in evaluation.

\begin{table}[t]
\centering
\caption{\textbf{Collected sources by published role.} The role is a property of the release rather than a
decision of ours, and what a source finally contributes is settled by the leakage check rather
than by the role alone.}
\label{tbl:roles}
\vspace{1mm}
\begin{NiceTabular}{l | r r | X[l]}[width=\linewidth]
\toprule
\textbf{Published role} & \textbf{Sources} & \textbf{Series} & \textbf{Sources in this group} \\
\midrule
\multirow{2}{*}{Pretrain only} & \multirow{2}{*}{8} & \multirow{2}{*}{8{,}542{,}433} & Time-300B~\citep{shi2024timemoe}, GIFT-Eval pretrain~\citep{aksu2024gifteval}, Chronos~\citep{ansari2024chronos}, LOTSA~\citep{woo2024unified}, EIA-930~\citep{eia930}, IMF PortWatch~\citep{portwatch}, NYC TLC~\citep{nyctlc}, NYC 311~\citep{nyc311} \\
\midrule
\multirow{3}{*}{No split} & \multirow{3}{*}{67} & \multirow{3}{*}{999{,}102} & Goiener smart meters~\citep{goiener}, UCI electricity load~\citep{ucielectricity}, Citi Bike~\citep{citibike}, Divvy~\citep{divvy}, NHS dispensing~\citep{nhsbsa}, Online Retail II~\citep{onlineretailii}, OPSD~\citep{opsd}, AEMO~\citep{aemo}, and 59 others \\
\midrule
\makecell[l]{Train only\\(Held-out private)} & 3 & 815{,}098 & Web Traffic~\citep{webtraffic}, Instacart~\citep{instacart}, Kaggle sales forecasting~\citep{kagglesales} \\
\midrule
\multirow{3}{*}{Train and test} & \multirow{3}{*}{11} & \multirow{3}{*}{50{,}489} & M4~\citep{makridakis2020m4}, M5~\citep{makridakis2022m5}, Favorita~\citep{favorita}, VN1~\citep{vn1}, Rossmann~\citep{rossmann}, Walmart sales forecast~\citep{walmartsales}, Walmart recruiting~\citep{walmartrecruiting}, Store Item Demand~\citep{storeitem}, Panama load (two releases)~\citep{panamaload}, Food demand~\citep{fooddemand} \\
\midrule
Evaluation & 3 & 899{,}618 & GIFT-Eval~\citep{aksu2024gifteval}, fev-bench~\citep{fevbench}, BOOM~\citep{boom} \\
\bottomrule
\end{NiceTabular}
\end{table}

\subsection{Preprocessing}
\label{sec:prep}

Preprocessing brings sources of all three types into a single schema. The demand filter of
\S\ref{sec:collect} applies to all of them, but at a different level in each. For competitions and
domain-specific open data (types 1 and 3) the source as a whole is demand, and the filter only
decides which columns to take. A general-purpose corpus (type 2) holds demand and non-demand
datasets side by side, and there the filter removes whole datasets. Of the seven general-purpose corpora in
Table~\ref{tbl:sources}, six are converted, since Monash is already contained in the Chronos and LOTSA
collections, and the filter keeps 269 of the 343 datasets they contain and drops the remaining 74. This large drop rate is expected, as these corpora are
assembled to cover \textit{time series} rather than \textit{demand}, and the dropped datasets are
mostly weather, generation, and air quality, which rules 1) and 2) of the demand filter in \S\ref{sec:collect} already exclude.

\textbf{General-purpose corpora.}
\label{sec:real_general}
These six are not single datasets but collections, each one containing many datasets under a single name, and the corpus itself therefore has no single domain. Four were published for pretraining
(Time-300B~\citep{shi2024timemoe}, the GIFT-Eval pretraining split~\citep{aksu2024gifteval}, the
Chronos training collection~\citep{ansari2024chronos}, LOTSA~\citep{woo2024unified}) and two as
evaluation suites (fev-bench~\citep{fevbench}, GIFT-Eval~\citep{aksu2024gifteval}). We therefore assign a
domain to each dataset inside them rather than to the corpus as a whole. These six hold 83\% of all series, and this one step therefore fixes the domain of most of the corpus.


\textbf{Conversion and verification.} All sources are converted to a single schema of
\texttt{(series\_id, timestamp, target)}, together with a per-series index recording the shape,
demand character, and provenance of each series. The converted counts are then verified against
numbers established \textit{outside} the pipeline, such as competition documentation or the
published number of entities, rather than against the conversion log itself.
Appendix~\ref{app:convert} records the schema in full and the four quantities that every source is
required to agree with before it is admitted to the corpus.

\begin{figure}[t]
\centering
\begin{minipage}[t]{0.63\linewidth}
\centering
\captionof{table}{\textbf{Statistics of demand datasets.} Counts and percentiles are over the whole
converted corpus, real-world and synthetic together.}
\label{tbl:corpus}
\vspace{1mm}
\begin{NiceTabular}{l r | l r}
\toprule
\textbf{[1] Quantity} & \textbf{Value} & \textbf{[2] Series length} & \textbf{Points} \\
\midrule
Series               & 11{,}306{,}740 & P25 & 128 \\
Observations         & 48{,}355{,}437{,}247 & P50 & 1{,}373 \\
Converted sources    & 73 & P75 & 8{,}761 \\
Collected sources    & 93 & P99 & 21{,}383 \\
\midrule
\textbf{[3] Class} & \textbf{Series} & \textbf{[4] Zero ratio} & \textbf{Share} \\
\midrule
Smooth       & 6{,}340{,}416 & P50 & 0.00 \\
Erratic      & 4{,}141{,}168 & P90 & 0.06 \\
Intermittent &   727{,}598   & P99 & 0.98 \\
Lumpy        &    97{,}558   & P100 & 1.00 \\
\bottomrule
\end{NiceTabular}
\end{minipage}
\hfill
\begin{minipage}[t]{0.34\linewidth}
\centering
\captionof{figure}{\textbf{Zeros in the corpus.} Zeros are absent from most series, and intermittency is therefore concentrated in a minority.}
\label{fig:corpus}
\vspace{1mm}
\includegraphics[width=\linewidth]{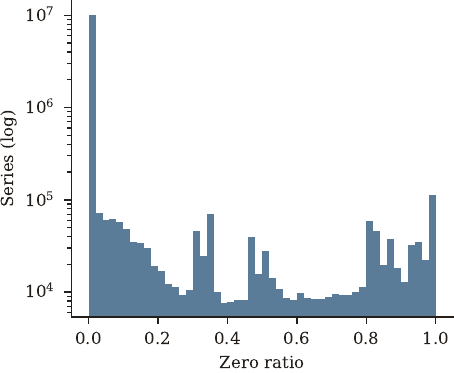}
\end{minipage}
\end{figure}

\textbf{Corpus statistics.}
The converted corpus contains 11.3M series and 48.4B observations from 73 sources. Aggregate size
alone does not indicate whether a corpus is suited to demand, as a small number of long
high-frequency series can dominate any total. Table~\ref{tbl:corpus} therefore reports the
\textit{shape} of the corpus instead, namely how long its series are, how they split across the four demand classes, and what share of
their values is zero at each percentile.

Two characteristics of the corpus affect all downstream design decisions. First, the length
distribution is heavily skewed, where the median series has 1{,}373 points while the lower quartile
has only 128. Second,
intermittency is concentrated, where zeros are absent from the median series but dominate the upper
percentiles, as shown in Figure~\ref{fig:corpus}. The intermittent and lumpy classes, the hard cases
that motivate the corpus, are therefore a minority by count, and the generator of
Section~\ref{sec:synth} is built to correct this imbalance by generating more series of the two rare classes.

\subsection{Train-Test Split}
\label{sec:split}

\textbf{Roles and their use.}
The five roles in Table~\ref{tbl:roles} reduce to a single rule, namely that only evaluation suites are held out, while all other roles are training candidates until
the leakage check below removes whatever overlaps the test side.

\begin{itemize}[topsep=1pt,itemsep=0.5pt,parsep=0pt]
\item \textbf{Evaluation} is the only role held out, as these three sources exist to be scored
against, and using any part of them for
training would invalidate the very measurement that they were published to provide.
\item \textbf{Pretrain only} and \textbf{No split} are used for training in full, as neither
withholds anything we are obliged to respect. What is removed from them is determined by the
leakage check below rather than by the role.
\item \textbf{Train only (held-out private)} is also used for training in full, since the hidden part was never published by its
authors, and there is therefore no test set of our own that any part of it could contaminate.
\item \textbf{Train and test} contributes \textit{both} parts to training, as its test part is the
competition's own rather than the benchmark we evaluate on, and any coincidence between the two
is caught by the leakage check below.
\end{itemize}

This rule concerns how a source was published rather than what it contains, and it is therefore not
sufficient on its own, since two sources can publish the same underlying series under different roles without any record of the overlap.

\textbf{Train and test separation.}
Evaluation uses the held-out evaluation suites, and all remaining series are available for
training. Separating them by dataset name is not sufficient, as the same underlying series is
redistributed under different names across sources. We therefore separate the two sides at the
level of the values themselves, where a redistributed copy cannot hide behind the name it was
republished under. Any training
candidate whose values match a test series is \textit{removed rather than down-weighted}. Table~\ref{tbl:split} reports what the check leaves on
each side, counted in series and in time points, together with what it removes from the corpus on the training side.

\begin{wraptable}{r}{0.60\linewidth}
\centering
\vspace{-4.3mm}
\caption{\textbf{Train and test separation.} Removed is what the leakage check takes out of
training, and the shares are over the whole corpus.}
\label{tbl:split}
\vspace{1mm}
\begin{NiceTabular}{l | r r | r r}
\toprule
& \multicolumn{2}{c|}{\textbf{Series}} & \multicolumn{2}{c}{\textbf{Time points}} \\
\cmidrule(lr){2-3} \cmidrule(lr){4-5}
& \textbf{Value} & \textbf{Share} & \textbf{Value} & \textbf{Share} \\
\midrule
Train   & 10{,}049{,}114 & 88.88\% & 47{,}419{,}572{,}862 & 98.06\% \\
Test    &    899{,}618 &  7.96\% &    810{,}403{,}280 &  1.68\% \\
Removed &    358{,}008 &  3.17\% &    125{,}461{,}105 &  0.26\% \\
\bottomrule
\end{NiceTabular}
\vspace{-2mm}
\end{wraptable}

We apply the same measurement to the pretraining data of the models we compare against, as discussed
in \S\ref{sec:exp_setup}. That data was fixed before the benchmarks and was never filtered against them, and the comparison is therefore \textit{not symmetric}, which we state rather than leave implicit. The
synthetic corpus is untouched by this check, since it holds no real observation that the leakage
check could match, and every one of its series is available to the training mixture in full.

\section{Synthetic Demand Dataset}
\label{sec:synth}

\begin{table}[!t]
\centering
\caption{\textbf{Design of the synthetic generator by demand properties.} Each property has a component of
its own in the sampling process, and each component is turned on or off for every series with its
strength drawn at random, so that the properties combine freely rather than in the fixed groups that
a fixed recipe would give.}
\label{tbl:synth}
\vspace{1mm}
\begin{NiceTabular}{l l | X[l]}[width=\linewidth]
\toprule
\textbf{Family} & \textbf{Property} & \textbf{Generating mechanism} \\
\midrule
\multirow{3}{*}{\makecell[l]{\textbf{Counts}}}
 & Discreteness        & Negative-binomial sampling from a multiplicative intensity \\
 & Overdispersion      & Dispersion parameter sampled per series \\
 & Intermittency       & Bernoulli mask over the intensity, targeted to a demand class \\
\midrule
\multirow{3}{*}{\makecell[l]{\textbf{Calendar}}}
 & Day-of-week profile & Multiplicative weekday factors \\
 & Seasonality         & Fourier components at annual and weekly periods \\
 & Moving holidays     & Lunar New Year, Chuseok, Easter, and Thanksgiving as shifting dates \\
\midrule
\multirow{3}{*}{\makecell[l]{\textbf{Promotion}}}
 & Uplift              & Three-phase response of build-up, peak, and post-promotion dip \\
 & Cannibalisation     & Negative cross-effect on related series \\
 & Price elasticity    & Log-linear response of intensity to a sampled price path \\
\midrule
\multirow{2}{*}{\makecell[l]{\textbf{Life-cycle}}}
 & Product launch      & Bass diffusion for adoption \\
 & Discontinuation     & Decaying tail with an absorbing zero state \\
\midrule
\multirow{2}{*}{\makecell[l]{\textbf{Observation}}}
 & Stock-out censoring & $(s,S)$ inventory policy clipping observed sales below latent demand \\
 & Reporting gaps      & Contiguous missing spans \\
\bottomrule
\end{NiceTabular}
\end{table}

The real-world corpus of Section~\ref{sec:data} is broad but uneven, since several of the behaviours
that decide whether a demand forecast is useful are thin in it, and one of them cannot appear in
observed data at all. We therefore generate a synthetic corpus alongside the real one, designed to cover exactly those
regions of demand behaviour.

\subsection{Motivation}
\label{sec:synth_why}

\textbf{Why synthetic data.}
Open demand data under-represents several behaviours that matter operationally, and in some cases
cannot represent them at all. Promotions and holidays are visible in the series only as unexplained level shifts. The calendar and the promotional plan that caused them are not published with the data. Product life-cycles are truncated, since a dataset released at one point in time rarely
covers both the launch and the discontinuation of the same item. Most importantly, observed sales are
a censored view of latent demand. When inventory runs out, a zero records an empty shelf rather than
an absence of customers, and no real dataset distinguishes the two.

We therefore generate a synthetic corpus of 47{,}500 series and 59.7M points alongside the real one.
Its purpose is not to substitute for real demand but to \textit{cover the regions of demand behaviour
that open data leaves thin}. The generating process is known, and it therefore records what observation alone cannot, as described in \S\ref{sec:synth_how}.

\subsection{Generating Process}
\label{sec:synth_how}

\begin{figure}[t]
\centering
\includegraphics[width=\linewidth]{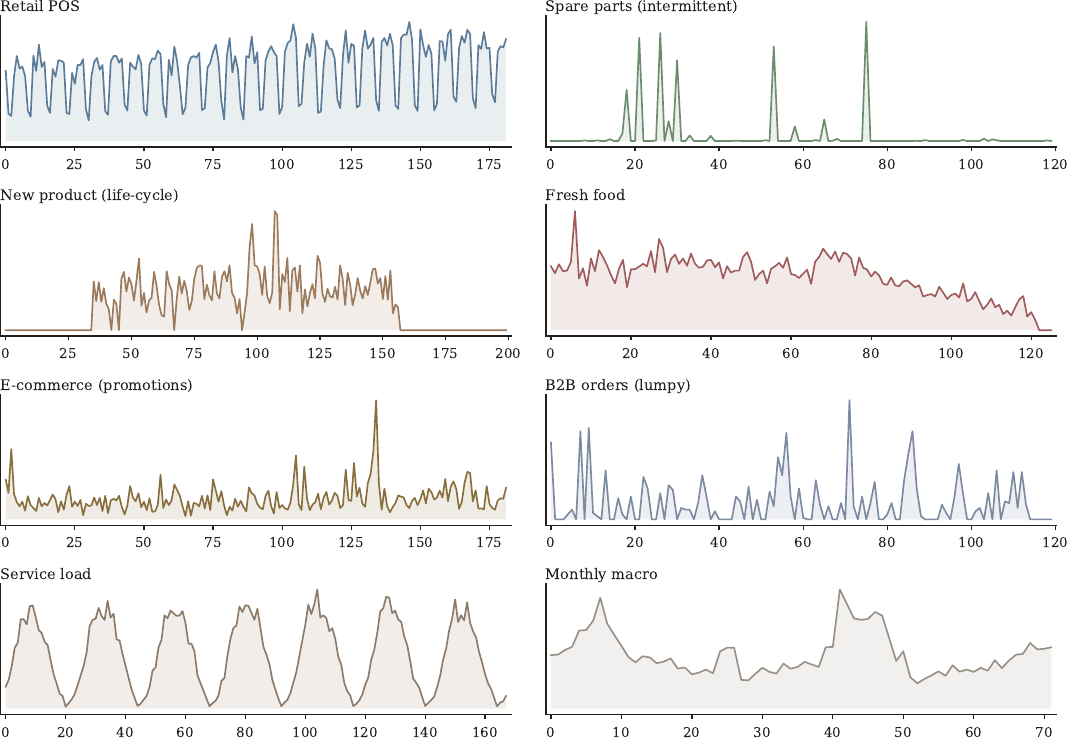}
\caption{\textbf{Representative synthetic demand series.} One series per generator. Spare parts and
B2B orders realise the intermittent and lumpy regimes that open data supplies only sparsely, new
product realises a full launch-to-discontinuation life-cycle, and e-commerce carries explicit promotional responses of its own.}
\label{fig:synth}
\end{figure}

\textbf{Generating process.}
A series is built in three stages, where a multiplicative intensity sets the expected level over
time, a count process turns that intensity into integer demand, and an observation model decides
what is actually recorded. Table~\ref{tbl:synth} lists the properties the generator reproduces and the mechanism
behind each. Every property is switched on independently with a sampled strength, which lets the corpus cover smooth seasonal demand and series dominated by long zero
runs and abrupt life-cycle transitions, together with the mixtures that lie between the two extremes.
The corpus is organised into eight generators, each configured to imitate a distinct operational
setting rather than a distinct statistical family. This keeps the corpus interpretable, since a
mixture can be described in terms of the business situations it emphasises.
Figure~\ref{fig:synth} shows one series from each of the eight generators below.

The corpus records the pre-censoring latent demand, the underlying intensity, and a stock-out flag
next to every observed value. This is a signal \textit{no real dataset can provide}, since in
observed sales data a zero caused by no customer and a zero caused by an empty shelf are
indistinguishable in the file as the publisher releases it.

\subsection{Validation and Use}
\label{sec:synth_valid}

\textbf{Validation.}
Generation is controlled rather than left to an unconstrained sampler, as the Syntetos--Boylan class is an \emph{input}
to the generator, which produces series to hit a target class, and the realised classification is
then checked against that target, where the two agree in 95.5\% of cases. Seven further validation
checks, covering class distribution, zero ratios, seasonality recovery, promotion response, censoring
rate, length distribution, and determinism under a fixed seed, all pass on the final corpus, which confirms that the properties that the generator was written to produce are the properties the corpus actually
carries rather than ones merely assumed from the generating process.

\textbf{Use in training.}
The synthetic corpus contains no real observations by construction, where no value of it can
match an evaluation series, and the leakage check therefore does not apply to it. It is therefore used in its entirety for
training, with no held-out portion. All 47{,}500 series are available to every mixture, and the train and test separation reported in \S\ref{sec:split} concerns the real-world corpus
alone, where a leaked series would otherwise be scored against itself.

\section{EXAONE-Demand}
\label{sec:moe}

\begin{figure}[t]
\centering
\includegraphics[width=\linewidth]{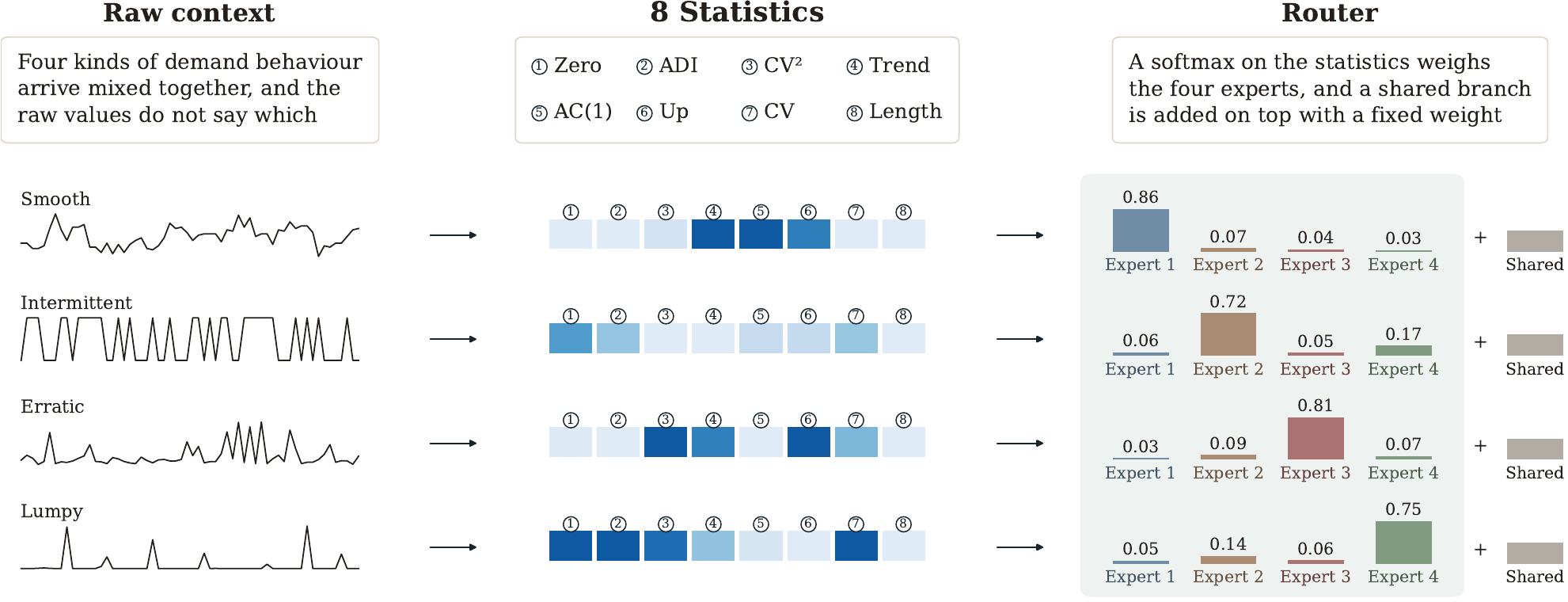}
\caption{\textbf{EXAONE Demand on the four demand classes.} (1) \textit{Raw context}: The window that the model receives, with no rescaling of
its values. (2) \textit{8 Statistics}: The scale-free summary that the window is reduced to.
(3) \textit{Router}: One weight per routed branch, with the always-on shared branch added at a
fixed weight.}
\label{fig:moe_arch}
\end{figure}

\textbf{Motivation.}
Demand series do not share one behaviour, since the corpus of Section~\ref{sec:data} spans
\textit{smooth} seasonal sales, \textit{intermittent} spare-part orders, \textit{erratic}
promotional spikes, and \textit{lumpy} wholesale batches. The hard cases are a minority by
count while being the reason the corpus exists, where one adapter fitted to all four spends
most of its capacity on the easy majority, and we therefore let it specialise by demand
behaviour rather than average across them.

\textbf{Overview of EXAONE Demand.}
EXAONE Demand keeps a pretrained TSFM frozen and adapts it with a mixture-of-experts (MoE)
adapter~\citep{shazeer2017outrageously, zadouri2024pushing}, whose five experts are low-rank
branches and whose mixing weights follow the demand behaviour of the input series. Four of them
are routed, each tied to one of the four demand classes, and the fifth is always on and shared by every series as a shared expert~\citep{dai2024deepseekmoe}, so that it carries the correction that every demand series
needs and leaves the routed four to carry only the differences between the classes.

The pipeline has five stages, and we describe each in turn:
\begin{itemize}[leftmargin=1.4em, itemsep=0.5pt, topsep=1pt]
  \item \textbf{Demand statistics} (\S\ref{sec:moe_stats}): We summarise the raw context window into eight scale-free statistics.
  \item \textbf{Continuous demand membership} (\S\ref{sec:moe_member}): We turn ADI and CV$^2$ into a membership over the four demand classes.
  \item \textbf{Routing} (\S\ref{sec:moe_route}): We route on the eight statistics and supervise the router with that membership.
  \item \textbf{Mixture of low-rank branches} (\S\ref{sec:moe_mix}): We mix the shared and the four routed branches in every frozen projection.
  \item \textbf{Training} (\S\ref{sec:moe_train}): We train only the low-rank branches and the router, keeping the backbone frozen.
\end{itemize}

Figure~\ref{fig:moe_arch} draws the whole pipeline, with four real series, one from
each of the four demand classes, and the statistics and the membership that each of them implies at the input of the adapter.

\subsection{Stage 1: Demand Statistics}
\label{sec:moe_stats}

The router reads eight statistics of the raw series, and each of them answers one question about the
demand behaviour that separates the four classes. We list them in the order they enter the router input:
\begin{itemize}[leftmargin=2.4em, itemsep=0.5pt, topsep=1pt]
  \item[{[1]}] \textbf{Zero share} (Zero): The fraction of the window with no demand, which is the plainest sign of intermittency.
  \item[{[2]}] \textbf{Average inter-demand interval} (ADI): The mean spacing between two non-zero periods, which is the first axis of the Syntetos--Boylan classification, separating smooth from intermittent demand.
  \item[{[3]}] \textbf{Squared coefficient of variation} (CV$^2$): The dispersion of the non-zero values around
        their own mean, which is the second axis of the same scheme, separating steady from erratic order sizes.
  \item[{[4]}] \textbf{Trend correlation} (Trend): The correlation between the series and time, which
        distinguishes a product in its launch or in its decline phase from one that has held a stable level.
  \item[{[5]}] \textbf{First-order autocorrelation} (AC(1)): How much one period predicts the
        next, which tells a seasonal or otherwise persistent series apart from one whose movements
        are close to noise.
  \item[{[6]}] \textbf{Fraction of upward steps} (Up): The share of consecutive pairs that increase, which
        describes the shape of the path the demand takes rather than its level or its spread.
  \item[{[7]}] \textbf{Coefficient of variation} (CV): The dispersion of the whole window, zeros
        included, which reacts to the spikes that the dispersion of the non-zero values alone would miss.
  \item[{[8]}] \textbf{Length} (Length): The number of observed periods, which says how much evidence the other seven rest on.
\end{itemize}

The first three are ratios over counts, the next three describe the shape of the path, and the last
two describe its spread and its size, and the eight together therefore cover both what the two classical
statistics of \S\ref{sec:moe_member} rest on and what they leave out. We now define each of them, from the raw window to the vector the router receives.

Let $x \in \mathbb{R}^{L}$ be the raw context window of one series and let $m \in \{0,1\}^{L}$ mark
its observed entries, with $n = \sum_t m_t$. We first decide which of those entries count as a
demand event, and because units differ by three orders of magnitude across the corpus, the test is relative to the
scale of the series rather than absolute,
\begin{equation}
\epsilon = 10^{-3} \cdot \frac{1}{n}\sum_t m_t \, |x_t| ,
\qquad
z_t = m_t \cdot \mathbb{1}\!\left[\,|x_t| > \epsilon\,\right] ,
\qquad
n_z = \sum_t z_t .
\end{equation}

From these we form the two axes of the Syntetos--Boylan classification~\citep{syntetos2005categorization},
namely the average inter-demand interval and the squared coefficient of variation of the non-zero
values,
\begin{equation}
\mathrm{ADI} = \frac{n}{n_z},
\qquad
\mu_z = \frac{1}{n_z}\sum_t z_t |x_t| ,
\qquad
\mathrm{CV}^2 = \frac{1}{n_z \mu_z^{2}}\sum_t z_t \left(|x_t| - \mu_z\right)^{2} .
\label{eq:adicv}
\end{equation}

Four further quantities describe the shape of the series. With $\bar{x}$ the mean over observed
entries, $\tilde{x}_t = m_t (x_t - \bar{x})$ the centred series, and $\tilde{t}$ the centred time
index, these are the trend correlation, the first-order autocorrelation, the fraction of upward
steps, and the overall coefficient of variation,
\begin{equation}
\rho_{\text{trend}} = \frac{\langle \tilde{t}, \tilde{x}\rangle}{\|\tilde{t}\|\,\|\tilde{x}\|},
\quad
\rho_{1} = \frac{\langle \tilde{x}_{1:L-1}, \tilde{x}_{2:L}\rangle}
                {\|\tilde{x}_{1:L-1}\|\,\|\tilde{x}_{2:L}\|},
\quad
u = \frac{1}{n-1}\sum_t \mathbb{1}\!\left[x_{t+1} > x_t\right],
\quad
\mathrm{CV} = \frac{\sigma}{|\bar{x}|} .
\end{equation}

The router input is the concatenation of eight numbers, where the three ratio-valued ones, namely
entries 2, 3, and 7, and the length in entry 8 are compressed on a logarithmic axis so that no
single statistic dominates the first layer,
\begin{equation}
s = \Big[\; 1 - \tfrac{n_z}{n},\;\;
\tfrac{1}{3}\log(1{+}\mathrm{ADI}),\;\;
\tfrac{1}{3}\log(1{+}\mathrm{CV}^2),\;\;
\rho_{\text{trend}},\;\;
\rho_{1},\;\;
2u - 1,\;\;
\tfrac{1}{3}\log(1{+}\mathrm{CV}),\;\;
\tfrac{1}{8}\log n \;\Big] \in \mathbb{R}^{8} .
\label{eq:stats}
\end{equation}
\begin{table}[t]
\centering
\caption{\textbf{Example router input for four demand series.} Each row is the vector $s$ of
Equation~\ref{eq:stats} as the router receives it, and the four series are drawn from four different
real-world sources and are of four different lengths.}
\label{tbl:stats}
\vspace{1mm}
\begin{NiceTabular}{l l | r r r r r r r r}
\toprule
\multirow{2.5}{*}{\textbf{Class}} & \multirow{2.5}{*}{\textbf{Source}} & \multicolumn{8}{c}{\textbf{8 Statistics}} \\
\cmidrule(lr){3-10}
 & & \textbf{Zero} & \textbf{ADI} & \textbf{CV$^2$} & \textbf{Trend}
& \textbf{AC(1)} & \textbf{Up} & \textbf{CV} & \textbf{Length} \\
\midrule
\textbf{Smooth}        & Citi Bike    & 0.00 & 0.23 & 0.04 & $-$0.16 & \phantom{$-$}0.36 & $-$0.87 & 0.10 & 0.56 \\
\textbf{Intermittent}  & Port calls   & 0.98 & 1.26 & 0.00 & $-$0.09 & $-$0.02 & $-$0.95 & 0.67 & 0.70 \\
\textbf{Erratic}       & SKU orders   & 0.03 & 0.24 & 0.32 & $-$0.04 & \phantom{$-$}0.10 & $-$0.66 & 0.28 & 0.51 \\
\textbf{Lumpy}         & Retail sales & 0.86 & 0.69 & 0.14 & $-$0.01 & \phantom{$-$}0.00 & $-$0.79 & 0.47 & 0.63 \\
\bottomrule
\end{NiceTabular}
\end{table}

Table~\ref{tbl:stats} shows $s$ for four series taken from the corpus, one from each of the four
demand classes. The classes separate on different entries, which is why eight numbers are kept
rather than the two that define them. The smooth and erratic series agree on the zero share and on
the interval, and are separated by the dispersion of their values, while the intermittent and lumpy
series are told apart by the interval and by that same dispersion together.

Every entry of $s$ is invariant to the unit of the series. A router that can see the magnitude of the values learns to recognise the \textit{source}
rather than the \textit{behaviour}, and a source it has not seen then falls outside everything it
has learned. The statistics are computed once per
series, before the encoder, and the same eight numbers are handed to every adapted layer, so that no
layer re-derives them from the hidden state that the backbone passes to it.

\subsection{Stage 2: Continuous Demand Membership}
\label{sec:moe_member}

\begin{wraptable}{r}{0.585\linewidth}
\centering
\vspace{-\intextsep}\vspace{-0.5mm}
\caption{\textbf{Classical demand classes.} A low ADI means that demand arrives often, and a high
CV$^2$ means that its size varies widely.}
\label{tbl:sbcuts}
\setlength{\tabcolsep}{4.5pt}
\begin{NiceTabular}{l | c c}
\toprule
& $\mathrm{CV}^2 < 0.49$ & $\mathrm{CV}^2 \geq 0.49$ \\
\midrule
$\mathrm{ADI} < 1.32$    & \makecell{\textbf{Smooth}\\Frequency\,$\uparrow$, Variability\,$\downarrow$}
                         & \makecell{\textbf{Erratic}\\Frequency\,$\uparrow$, Variability\,$\uparrow$} \\
\midrule
$\mathrm{ADI} \geq 1.32$ & \makecell{\textbf{Intermittent}\\Frequency\,$\downarrow$, Variability\,$\downarrow$}
                         & \makecell{\textbf{Lumpy}\\Frequency\,$\downarrow$, Variability\,$\uparrow$} \\
\bottomrule
\end{NiceTabular}
\vspace{-2mm}
\end{wraptable}

The classical scheme of the intermittent demand literature~\citep{croston1972forecasting,
syntetos2001accuracy, syntetos2005categorization} assigns a series to one of four classes by cutting
Equation~\ref{eq:adicv} at $\mathrm{ADI} = 1.32$ and $\mathrm{CV}^2 = 0.49$, as shown in
Table~\ref{tbl:sbcuts}. However, demand series do not respect those cuts. They spread \textit{continuously} across the two axes, and a
series just past a boundary is not a different kind of object from the one that lies just before it. Series at ADI $= 1.30$ and $1.34$ are nearly alike, yet the cuts split them.

We therefore keep the two axes but soften the cuts, and since both quantities are ratios, we measure the
distance to each cut on a logarithmic axis, where the gap between $\mathrm{ADI} = 1.2$ and $1.4$
carries the same weight as the gap between $8.0$ and $9.3$,
\begin{equation}
a = \sigma\!\left(\frac{\log \mathrm{ADI} - \log 1.32}{\tau_a}\right),
\qquad
c = \sigma\!\left(\frac{\log \mathrm{CV}^2 - \log 0.49}{\tau_c}\right),
\label{eq:softcut}
\end{equation}
with $\sigma$ the logistic function and $\tau_a, \tau_c$ the widths of the cuts. Here $a$ reads as how
\textit{intermittent} the series is and $c$ as how \textit{erratic}, and treating the two axes as independent gives a
membership whose entries sum to one,
\begin{equation}
\pi = \big[\,(1-a)(1-c),\;\; a(1-c),\;\; (1-a)c,\;\; a\,c\,\big]
\;=\; [\,\pi_{\text{smooth}},\; \pi_{\text{inter}},\; \pi_{\text{erratic}},\; \pi_{\text{lumpy}}\,].
\label{eq:member}
\end{equation}

A series that is clearly one thing gets a membership close to a corner, while a series between two
behaviours keeps \textit{both} of them. The lumpy example of
Table~\ref{tbl:stats} has $\pi = (0.00, 0.46, 0.00, 0.53)$, which says it is lumpy by a narrow
margin over intermittent, a split between the two classes that a hard label would have thrown away.

\begin{figure}[t]
\centering
\begin{minipage}[t]{0.372\linewidth}
\centering
\vspace{0pt}
\includegraphics[width=\linewidth]{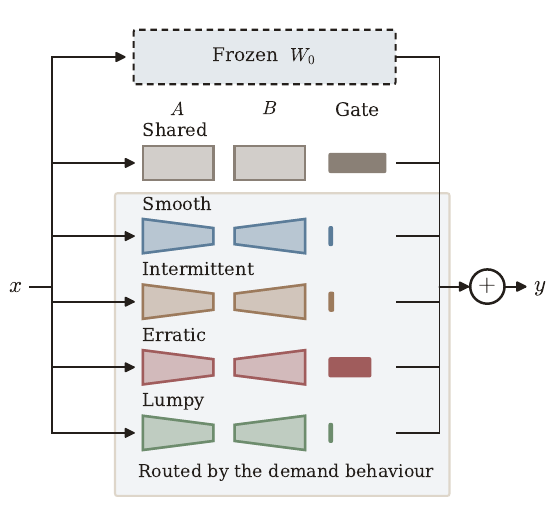}
\caption{\textbf{One adapted projection.} The four routed branches are weighed by the gate, and the
shared branch is always on.}
\label{fig:molora}
\end{minipage}\hfill
\begin{minipage}[t]{0.588\linewidth}
\centering
\vspace{0pt}
\includegraphics[width=\linewidth]{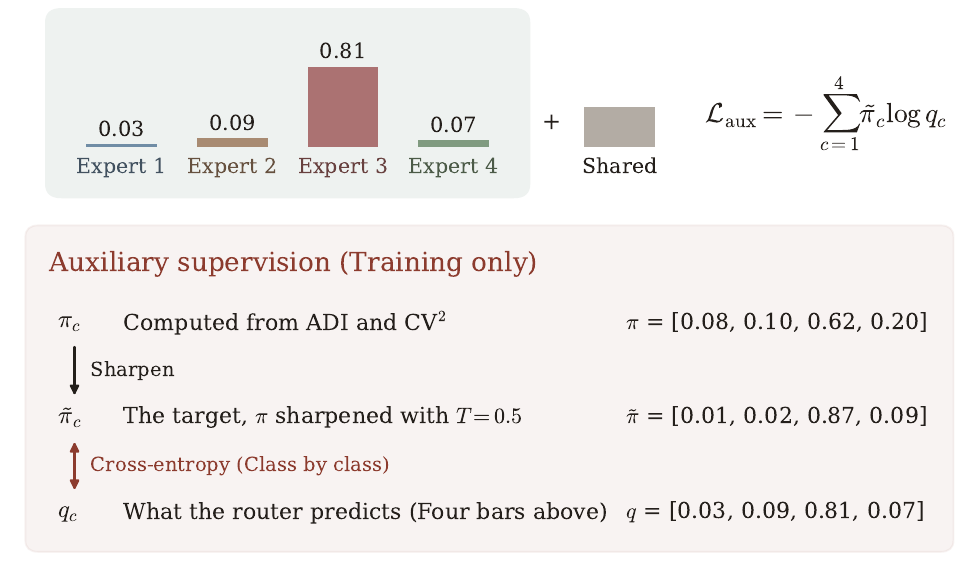}
\caption{\textbf{Auxiliary supervision on one erratic series.} The bars are the router output, and
the chain below turns the membership into the target that the cross-entropy pulls that output
towards during training.}
\label{fig:moe_aux}
\end{minipage}
\end{figure}

\subsection{Stage 3: Routing}
\label{sec:moe_route}

Each of the four routed branches is meant to carry one demand behaviour, and a rule has to decide
how much of each branch a given series receives, which makes the membership of
Equation~\ref{eq:member} the obvious candidate for that rule. It rests on two of the eight
statistics, however, and has to be recomputed from ADI and CV$^2$ for every series that arrives.
Therefore, we learn that rule instead, as a small network that reads the whole vector $s$ and that we
call the \textit{router}, while the membership is kept as a supervision signal rather than as the
mixture itself. The auxiliary term of \S\ref{sec:moe_train} makes that arrangement
precise, and it is that term which ties each routed branch to one demand class.

That router is a two-layer network applied to $s$, followed by a softmax,
\begin{equation}
q = \mathrm{softmax}\!\left( \frac{W_2 \,\phi(W_1 s + b_1)}{\tau} \right) \in \Delta^{3},
\qquad W_1 \in \mathbb{R}^{h \times 8}, \;\; W_2 \in \mathbb{R}^{4 \times h},
\label{eq:gate}
\end{equation}
where $\phi$ is a GELU nonlinearity, $\tau$ is a temperature, and $h$ is the hidden width. The router produces four weights,
one per demand class of Equation~\ref{eq:member}. The shared branch is not routed at all, as
it takes a fixed weight $w$ while the routed branches divide whatever remains,
\begin{equation}
g = \big[\, w, \;\; (1-w)\,q_1, \;\; (1-w)\,q_2, \;\; (1-w)\,q_3, \;\; (1-w)\,q_4 \,\big]
\in \Delta^{4} .
\label{eq:share}
\end{equation}
One gate is produced per series rather than per token, and every token of a series is adapted by the
same mixture.

Figure~\ref{fig:molora} draws what one adapted projection then holds, namely the frozen weight, the
shared branch, and the four routed ones that the gate weighs. Each adapted layer carries its own router, and every one of them reads the same vector $s$ and decides its own mixture from it. The router is small on purpose and cannot learn anything beyond a partition of the eight-dimensional
space of statistics, which is all the design asks of it and all the supervision teaches.

\begin{figure}[t]
\centering
\includegraphics[width=\linewidth]{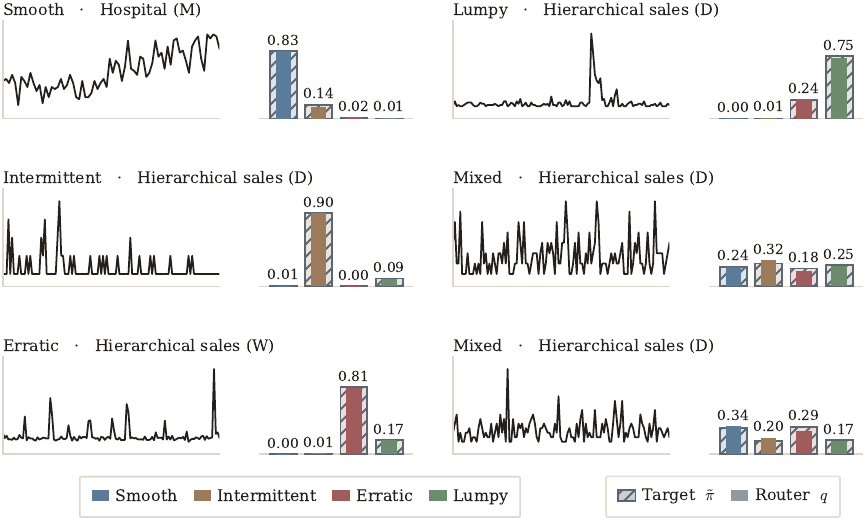}
\caption{\textbf{What the router produces on real demand series.} Each panel holds a context window
and the weight that the gate gives to each of the four routed branches, against the target that
supervision asks for, where the four bars carry the smooth, intermittent, erratic, and lumpy
branches of the mixture in that same order.}
\label{fig:router}
\end{figure}

Nothing in Equation~\ref{eq:gate} ties routed branch $c$ to any particular behaviour. Left alone the
routed branches are \textit{interchangeable}, and whichever partition the gate settles on is an \textit{accident
of initialisation}. We therefore supervise the gate with the membership of
Equation~\ref{eq:member}, using a cross-entropy between the two distributions,
\begin{equation}
\mathcal{L} = \mathcal{L}_{\text{forecast}} \;+\; \lambda \cdot \mathcal{L}_{\text{aux}},
\qquad
\mathcal{L}_{\text{aux}} = -\sum_{c=1}^{4} \tilde{\pi}_c \log q_c .
\label{eq:aux}
\end{equation}

The loss is written on $q$ rather than on $g$, which leaves the shared branch out of it and
supervises only the four routed branches. That is what gives routed branch $c$ a meaning of its own,
in that branch 1 is pulled toward smooth series and branch 2 toward intermittent ones, while a
series that is $0.6$ lumpy and $0.4$ intermittent asks for both in that proportion.

The target $\tilde{\pi}$ is a sharpened form of the membership of Equation~\ref{eq:member},
\begin{equation}
\tilde{\pi}_c \;=\; \frac{\pi_c^{1/T}}{\sum_{c'} \pi_{c'}^{1/T}} ,
\label{eq:sharp}
\end{equation}
where $T < 1$ is the sharpening temperature. We sharpen because series near the cuts of Equation~\ref{eq:adicv} carry a membership that is close
to a tie, and a target such as $[0, 0.40, 0, 0.60]$ is too weak to settle which branch should lead.
The alternative is a hard label and a negative log-likelihood on the
assigned class, which pulls $q$ toward a one-hot vector. We do not use it, because it contradicts
the premise of Equation~\ref{eq:member}, namely that a demand series is a \textit{mixture of
behaviours} rather than a \textit{member of one class}.

Figure~\ref{fig:moe_aux} follows one erratic series through that chain, from the membership we
measure to the target we sharpen and the weights the router finally produces. The auxiliary term is
a \textit{training signal only}, and at inference the router reads the eight statistics of the raw window and no
membership is computed, as only the loss needs it.

Figure~\ref{fig:router} puts the trained router of EXAONE Demand on six real series to show
what those weights look like once training is done. The four panels named by a class are the series
on which the router is the most decisive, and on them the weight it produces sits almost exactly on
the target that the supervision asks for. The two panels named \textit{Mixed} are the series on
which it is the least decisive, and the target there is spread across three branches while the
router spreads its weight in the same way, which is what a series that mixes two behaviours is
meant to look like at the gate.

\subsection{Stage 4: Mixture of Low-Rank Branches}
\label{sec:moe_mix}

We replace six projections in each encoder block, namely the four attention projections and the two
feed-forward projections, and leave the rest of the block untouched. Each adapted layer keeps its
frozen weight $W_0$ and adds $E = 5$ low-rank branches mixed by $g$,
\begin{equation}
y \;=\; W_0\,x \;+\; \sum_{e=0}^{4} g_e \cdot \frac{\alpha}{r_e} \cdot B_e A_e\, x ,
\qquad A_e \in \mathbb{R}^{r_e \times d_{\text{in}}}, \;\;
B_e \in \mathbb{R}^{d_{\text{out}} \times r_e} .
\label{eq:molora}
\end{equation}
Branch 0 is the shared one and branches 1 to 4 are the routed ones. The shared branch is given the larger rank, so that it alone holds as much capacity as a plain
low-rank adapter, while each routed branch adds a small correction on top of it. A shared branch that is too small would
leave part of the common correction to the routed branches, and that residue would blur the routing
that the supervision of Stage~3 asks for. The scale $\alpha / r_e$ follows the low-rank convention and is applied per branch, so that a
change of rank does not change the scale at which a branch enters the sum.

Written naively, Equation~\ref{eq:molora} forms one output tensor per branch. We instead stack the
branches along the rank axis and evaluate the sum with two matrix products,
\begin{equation}
y = W_0 x + \bar{B}\,\big( (\bar{A} x) \odot (g \otimes \mathbb{1}_r) \big),
\qquad
\bar{A} = \begin{bmatrix} A_0 \\ \vdots \\ A_4 \end{bmatrix},
\quad
\bar{B} = \begin{bmatrix} B_0 & \cdots & B_4 \end{bmatrix},
\end{equation}
where the only intermediate tensor has width $\sum_e r_e$ rather than $E \, d_{\text{out}}$.

\subsection{Stage 5: Training}
\label{sec:moe_train}

Only $A_e$, $B_e$, and the router of Equation~\ref{eq:gate} receive gradients. Everything else stays
frozen, including the embedding and the normalisation layers. The loss is Equation~\ref{eq:aux}, and the weight $\lambda$ is held fixed for every run that we report.

The three choices below are what keep the four routed branches apart from each other, and each of
them does so in a different way, namely how the router starts, what it reads, and how much weight the
shared branch is given.

\textbf{Router starts at random.} With $W_2 = 0$ the softmax of Equation~\ref{eq:gate} is
exactly uniform for every input, every routed branch receives the same gradient, and the symmetry between
them never breaks. A fixed uniform mixture of $E$ branches of rank $r$ equals a single adapter of
rank $E r$ by Equation~\ref{eq:molora}, and the architecture then reduces to the thing it was meant to
improve on. Drawing $W_2$ at random costs nothing, since the adapter output is multiplied by $B_e = 0$ at step
zero and the adapted layer therefore returns $W_0 x$ exactly, no matter which weights the router
assigns.

\textbf{Router reads the raw series rather than the hidden state.} A gate applied to hidden
states learns to recognise something the backbone encodes for itself, and its softmax flattens for
the same reason as above. The raw series gives the router information that the encoder
\textit{never sees}, since the encoder is fed a normalised window.

\textbf{Shared weight balances the two parts.} The weight $w$ of Equation~\ref{eq:share}
decides how much of the adapter is common and how much is routed. Near zero the routed branches
carry almost everything and near one they carry almost nothing, and we therefore set $w$ between the
two, where both parts contribute, at $w = 0.5$ in every run that we report.


\section{Experiments}
\label{sec:exp}

\subsection{Datasets and Metrics}
\label{sec:exp_setup}

\textbf{Evaluation suite.} We evaluate on 22 held-out datasets, none of which takes any part in
training, as verified by the leakage check of Section~\ref{sec:data} at the level of \textit{raw
values} rather than \textit{dataset names}. Every model is scored on the same window, the same horizon, and
the same quantile levels, with no per-model tuning of any kind.

\textbf{Metrics.} We report the six metrics of Table~\ref{tbl:main}, namely the mean absolute
scaled error (MASE), the normalised deviation (ND), the weighted quantile loss (WQL), the mean
absolute percentage error (MAPE), the mean absolute error (MAE), and the mean scaled interval score
(MSIS), and we aggregate every one of them across the 22 datasets by geometric mean, which gives
every dataset the same share of the aggregate whatever level its errors sit at.

Let $y_t$ be the observed value at step $t$ of a horizon of length $h$, let
$\hat{y}_t$ be the median forecast, let $\hat{y}^{(q)}_t$ be the forecast at quantile level $q$,
and let $x_1, \dots, x_n$ be the training portion of the same series at seasonal period $m$, which
the first and the last of the six use alone. The six of them are written as follows,
\begin{equation}
\mathrm{MASE} = \frac{\frac{1}{h}\sum_{t=1}^{h} \lvert y_t - \hat{y}_t \rvert}
                      {\frac{1}{n-m}\sum_{t=m+1}^{n} \lvert x_t - x_{t-m} \rvert},
\qquad
\mathrm{ND} = \frac{\sum_{t=1}^{h} \lvert y_t - \hat{y}_t \rvert}
                    {\sum_{t=1}^{h} \lvert y_t \rvert},
\label{eq:mase}
\end{equation}
\vspace{2.5mm}
\begin{equation}
\mathrm{WQL} = \frac{\sum_{q \in Q} \sum_{t=1}^{h}
       2\,\big(q\,[\,y_t - \hat{y}^{(q)}_t\,]_{+} + (1-q)\,[\,\hat{y}^{(q)}_t - y_t\,]_{+}\big)}
      {\lvert Q \rvert \sum_{t=1}^{h} \lvert y_t \rvert},
\qquad
\mathrm{MAPE} = \frac{1}{h}\sum_{t=1}^{h}
       \frac{\lvert y_t - \hat{y}_t \rvert}{\lvert y_t \rvert},
\label{eq:wql}
\end{equation}
\vspace{2.5mm}
\begin{equation}
\mathrm{MAE} = \frac{1}{h}\sum_{t=1}^{h} \lvert y_t - \hat{y}_t \rvert,
\qquad
\mathrm{MSIS} = \frac{\frac{1}{h}\sum_{t=1}^{h}
   \Big( u_t - l_t + \tfrac{2}{\alpha}\big[\,l_t - y_t\,\big]_{+}
                    + \tfrac{2}{\alpha}\big[\,y_t - u_t\,\big]_{+} \Big)}
        {\frac{1}{n-m}\sum_{t=m+1}^{n} \lvert x_t - x_{t-m} \rvert},
\label{eq:msis}
\end{equation}
\vspace{0mm}

\noindent where $Q$ is the set of nine quantile levels that both of our models emit, $[\,\cdot\,]_{+}$ keeps
the positive part of its argument, and $l_t$ and $u_t$ are the bounds of the central $1 - \alpha$
interval at $\alpha = 0.2$. Each of the six divides the error by a different quantity, and that divisor is what the metric is about.
All six are non-negative and unbounded above, and lower is better on each:
\begin{itemize}[leftmargin=1.4em, itemsep=0.5pt, topsep=1pt]
  \item \textbf{\textit{MASE}}: How much better than doing nothing a model is, since the divisor is the
        error that a seasonal naive forecaster makes on the history of the same series. A value
        above one says that repeating last season would have been the better choice of the two,
        and every row of Table~\ref{tbl:main} sits above one on the present evaluation suite.
  \item \textbf{\textit{ND}}: The share of the demand that the forecast misplaces, since the divisor is
        the total volume of the horizon. A value of $0.14$ reads as fourteen units misplaced for
        every hundred demanded, the currency a planner stocks in.
  \item \textbf{\textit{WQL}}: How good the whole predictive distribution is rather than its middle
        alone, since the error at each quantile level is charged asymmetrically. Missing above the
        $0.9$ quantile costs nine times as much as missing below it.
  \item \textbf{\textit{MAPE}}: The error as a fraction of the demand at each step taken separately,
        which is the most familiar of the six and the least suited to this setting. A value of
        $0.26$ reads as a quarter of the demand missed on the average period.
  \item \textbf{\textit{MAE}}: The average absolute error, and the only one of the six that keeps the
        unit of the series. Its level says nothing across datasets, while its ordering across the
        models of one suite still says what the others say.
  \item \textbf{\textit{MSIS}}: How good the interval is rather than the point or the distribution,
        charging the width of the central $80\%$ band plus a penalty for every observation outside
        it. It is the one metric that widening the band alone cannot win.
\end{itemize}

\textbf{Average rank and win rate.} To separate \textit{how much} a model is ahead from
\textit{how often} it is ahead, we report two further quantities beside the six metrics, both of
which read the per-dataset MASE rather than the aggregate of it. An aggregate is a single number
that a handful of datasets can carry, and a model that is far ahead on three datasets and behind on
the other nineteen can lead on it, which is the reading that these two are there to rule out.

The \textit{average rank} of a model is its mean position across the 22 datasets, where a model
that is second on every one of them scores 2 however wide or narrow the gaps around it are.
The \textit{win rate} is the share of the pairwise comparisons that a model wins, and with 38 other
rows and 22 datasets there are 836 comparisons for each of them. A model wins one of those when its
MASE on that dataset is the lower of the two, and a tie counts as a win for neither model.

\subsection{Experimental Setup}
\label{sec:exp_impl}

\textbf{Baselines.} We compare against the 36 TSFMs of Table~\ref{tbl:main} and the frozen backbone
that we adapt, all \textit{run by us} rather than
\textit{quoted from their papers}. The comparison covers Chronos-Bolt (Base, Small) and Chronos-T5 (Base,
Small)~\citep{ansari2024chronos}, Chronos-2 and Chronos-2 (Synthetic)~\citep{ansari2025chronos2},
TiRex-1.1~\citep{auer2025tirex}, TimesFM-1.0, TimesFM-2.0 and TimesFM-2.5~\citep{das2024decoder}, Moirai-1.1-R (Small, Base, Large) and Moirai-2.0-R~\citep{woo2024unified}, Timer-S1~\citep{liu2024timer}, Sundial~\citep{liu2025sundial}, Toto (Open Base)~\citep{cohen2024toto},
TabPFN-TS~\citep{hoo2025tabpfnts}, FlowState and Granite FlowState-R1~\citep{graf2026flowstate},
TTM-R1 and TTM-R2~\citep{ekambaram2024ttm}, PatchTST-FM-R1 and Granite
PatchTST-FM-R1~\citep{nie2023patchtst}, Kairos (10M, 23M, 50M)~\citep{feng2025kairos},
TempoPFN~\citep{moroshan2025tempopfn}, Reverso~\citep{fu2026reverso},
VisionTS~\citep{chen2025visionts}, Lag-Llama~\citep{rasul2023lagllama}, CleanTS~\citep{eink2025cleants}, and YingLong (6M, 50M, 110M, 300M)~\citep{wang2025yinglong}. No baseline is selected by how well it does on the 22 datasets, and every model, ours included, is scored
under the same protocol, with no chance to tune on it at any stage of the study that this section reports and the appendix records in full.


\textbf{Implementation.} The adapter of Section~\ref{sec:moe} is attached to the four attention
projections and the two feed-forward projections of every encoder block, and the backbone stays
frozen throughout. The shared branch is given the larger rank and half of the mixture weight, and
the router is trained with the auxiliary term of Equation~\ref{eq:aux} alongside the forecast loss.
Appendix~\ref{app:hparams} records every hyperparameter behind the
tables and figures of this section.

\subsection{Two Versions of EXAONE Demand}
\label{sec:exp_release}

\begin{table}[t]
\centering
\caption{\textbf{Comparison with TSFMs.} Both of our models are shown against every baseline we
ran and against the frozen backbone that we adapt, on the 22 evaluation datasets, with the rows
ordered and numbered (\#) by MASE.}
\label{tbl:main}
\vspace{1mm}
\setlength{\tabcolsep}{3.5pt}
\begin{tabular}{c l !{\color{black}\vrule} c c !{\color{black}\vrule} c c c c c c}
\toprule
\multirow{2.5}{*}{\textbf{\#}} & \multirow{2.5}{*}{\textbf{Model}} & \multicolumn{2}{c|}{\textbf{Ranking}} & \multicolumn{6}{c}{\textbf{Error metrics}} \\
\cmidrule(lr){3-4} \cmidrule(lr){5-10}
 & & \textbf{Avg. rank} & \textbf{Win rate} & \textbf{MASE} & \textbf{ND} & \textbf{WQL} & \textbf{MAPE} & \textbf{MAE} & \textbf{MSIS} \\
\midrule
\rowcolor{LightPurple} \textbf{1} & EXAONE Demand             & \best{4.09} & \best{91.9\%} & \best{1.0667} & \best{0.1409} & \best{0.1138} & \best{0.2565} & \best{60.24} & \second{10.70} \\
\rowcolor{LightPurple} \textbf{2} & EXAONE Demand (Synthetic) & \second{5.55} & \second{88.0\%} & \second{1.0742} & \second{0.1420} & \second{0.1147} & \second{0.2583} & \second{60.73} & \best{10.63} \\
\midrule
3 & TiRex-1.1~\citep{auer2025tirex}                  & 7.64 & 82.5\% & 1.0818 & 0.1451 & 0.1161 & 0.2911 & 62.06 & 10.75 \\
4 & Chronos-2~\citep{ansari2025chronos2}             & 8.59 & 80.0\% & 1.0885 & 0.1497 & 0.1406 & 0.3010 & 64.02 & 14.48 \\
5 & EXAONE Backbone (Zero-shot)                        & 7.95 & 81.7\% & 1.1050 & 0.1445 & 0.1169 & 0.2609 & 61.82 & 11.41 \\
6 & TimesFM-2.5~\citep{das2024decoder}               & 10.05 & 76.2\% & 1.1073 & 0.1486 & 0.1215 & 0.3032 & 63.58 & 12.31 \\
7 & Chronos-Bolt (Base)~\citep{ansari2024chronos}    & 11.05 & 73.6\% & 1.1122 & 0.1493 & 0.1191 & 0.2941 & 63.84 & 11.14 \\
8 & Timer-S1~\citep{liu2024timer}                    & 12.27 & 70.3\% & 1.1197 & 0.1468 & 0.1171 & 0.2997 & 62.77 & 12.08 \\
9 & Reverso~\citep{fu2026reverso}                    & 11.27 & 73.0\% & 1.1291 & 0.1510 & 0.1510 & 0.2773 & 64.60 & 45.17 \\
10 & Chronos-Bolt (Small)~\citep{ansari2024chronos}   & 14.00 & 65.8\% & 1.1331 & 0.1485 & 0.1193 & 0.2998 & 63.53 & 11.57 \\
11 & Chronos-T5 (Base)~\citep{ansari2024chronos}      & 8.68 & 79.8\% & 1.1336 & 0.1494 & 0.1219 & 0.2712 & 63.90 & 13.33 \\
12 & Toto (Open Base)~\citep{cohen2024toto}           & 11.73 & 71.8\% & 1.1399 & 0.1495 & 0.1214 & 0.2677 & 63.93 & 11.64 \\
13 & Chronos-2 (Synthetic)~\citep{ansari2025chronos2} & 15.95 & 60.6\% & 1.1515 & 0.1594 & 0.1452 & 0.3088 & 68.17 & 16.11 \\
14 & Moirai-2.0-R~\citep{woo2024unified}              & 12.05 & 70.9\% & 1.1586 & 0.1531 & 0.1260 & 0.2802 & 65.48 & 13.56 \\
15 & Chronos-T5 (Small)~\citep{ansari2024chronos}     & 12.27 & 70.3\% & 1.1623 & 0.1524 & 0.1249 & 0.2751 & 65.18 & 13.87 \\
16 & Sundial~\citep{liu2025sundial}                   & 15.68 & 61.4\% & 1.1690 & 0.1543 & 0.1325 & 0.3006 & 66.00 & 19.26 \\
17 & TabPFN-TS~\citep{hoo2025tabpfnts}                & 20.23 & 49.4\% & 1.2172 & 0.1719 & 0.1353 & 0.3520 & 73.52 & 12.21 \\
18 & FlowState~\citep{graf2026flowstate}              & 19.23 & 52.0\% & 1.2363 & 0.1705 & 0.1357 & 0.3422 & 72.94 & 12.14 \\
19 & Granite FlowState-R1~\citep{graf2026flowstate}   & 18.68 & 53.5\% & 1.2419 & 0.1710 & 0.1363 & 0.3469 & 73.12 & 12.67 \\
20 & Moirai-1.1-R (Large)~\citep{woo2024unified}      & 18.23 & 54.7\% & 1.2572 & 0.1640 & 0.1322 & 0.3027 & 70.16 & 12.51 \\
21 & Moirai-1.1-R (Base)~\citep{woo2024unified}       & 19.09 & 52.4\% & 1.2724 & 0.1638 & 0.1331 & 0.3108 & 70.06 & 12.67 \\
22 & TimesFM-1.0~\citep{das2024decoder}               & 20.05 & 49.9\% & 1.2794 & 0.1727 & 0.1427 & 0.3448 & 73.87 & 16.27 \\
23 & PatchTST-FM-R1~\citep{nie2023patchtst}           & 21.23 & 46.8\% & 1.2984 & 0.1693 & 0.1384 & 0.3312 & 72.41 & 12.91 \\
24 & Moirai-1.1-R (Small)~\citep{woo2024unified}      & 22.50 & 43.4\% & 1.3570 & 0.1753 & 0.1423 & 0.3230 & 74.98 & 13.58 \\
25 & TempoPFN~\citep{moroshan2025tempopfn}            & 21.68 & 45.6\% & 1.3968 & 0.1794 & 0.1475 & 0.3716 & 76.73 & 19.61 \\
26 & Granite PatchTST-FM-R1~\citep{nie2023patchtst}   & 23.41 & 41.0\% & 1.4061 & 0.1790 & 0.1493 & 0.3481 & 76.54 & 16.25 \\
27 & Kairos (50M)~\citep{feng2025kairos}              & 22.23 & 44.1\% & 1.4641 & 0.1905 & 0.1572 & 0.3648 & 81.46 & 18.97 \\
28 & Kairos (23M)~\citep{feng2025kairos}              & 26.00 & 34.2\% & 1.5391 & 0.2081 & 0.1737 & 0.3730 & 89.02 & 20.87 \\
29 & TTM-R1~\citep{ekambaram2024ttm}                  & 27.91 & 29.2\% & 1.5714 & 0.2014 & 0.2014 & 0.4047 & 86.13 & 62.86 \\
30 & TTM-R2~\citep{ekambaram2024ttm}                  & 29.73 & 24.4\% & 1.6399 & 0.2125 & 0.2125 & 0.4387 & 90.89 & 65.60 \\
31 & Kairos (10M)~\citep{feng2025kairos}              & 26.91 & 31.8\% & 1.6463 & 0.2067 & 0.1705 & 0.3909 & 88.42 & 21.05 \\
32 & VisionTS~\citep{chen2025visionts}                & 30.27 & 23.0\% & 1.9069 & 0.2444 & 0.2444 & 0.5455 & 104.55 & 76.28 \\
33 & Lag-Llama~\citep{rasul2023lagllama}              & 29.73 & 24.4\% & 2.0328 & 0.2437 & 0.2078 & 0.4339 & 104.24 & 24.97 \\
34 & TimesFM-2.0~\citep{das2024decoder}               & 33.14 & 15.4\% & 2.4684 & 0.2548 & 0.1989 & 0.4632 & 108.98 & 20.09 \\
35 & CleanTS~\citep{eink2025cleants}                  & 33.86 & 13.5\% & 3.1566 & 0.3705 & 0.3112 & 0.7209 & 158.48 & 45.16 \\
36 & YingLong (6M)~\citep{wang2025yinglong}           & 36.00 & 7.9\% & 3.7329 & 0.4672 & 0.4051 & 0.8661 & 199.83 & 73.96 \\
37 & YingLong (110M)~\citep{wang2025yinglong}         & 36.82 & 5.7\% & 3.8443 & 0.4773 & 0.4289 & 0.8865 & 204.14 & 88.79 \\
38 & YingLong (300M)~\citep{wang2025yinglong}         & 36.86 & 5.6\% & 3.8491 & 0.4879 & 0.4319 & 0.8870 & 208.69 & 85.43 \\
39 & YingLong (50M)~\citep{wang2025yinglong}          & 37.41 & 4.2\% & 3.8853 & 0.4861 & 0.4348 & 0.8876 & 207.93 & 87.12 \\
\bottomrule
\end{tabular}
\end{table}

We build two models that differ only in their training data. \textbf{EXAONE Demand}, the model of
Section~\ref{sec:moe}, is trained on real-world and synthetic demand together, and \textbf{EXAONE
Demand (Synthetic)} on the synthetic corpus alone, with no real-world series taking part at any
point of training. The second model exists because open demand data carries licences that a model trained on it
inherits, while one trained on series we generated \textit{ourselves} inherits none of them. The
question is whether the synthetic corpus can \textit{stand on its own} rather than only
\textit{fill gaps}.

\subsection{Main Results}
\label{sec:exp_main}

Table~\ref{tbl:main} reports both of our models against every baseline we ran, on the 22
evaluation datasets and under the protocol of \S\ref{sec:exp_setup}. Both of them lead the table on
every one of the six metrics as well as on the average rank and the win rate, and neither was tuned
against the suite that scores them. The two take the first and the second place on five of the six
metrics, and they divide those two places between themselves on MSIS, where the synthetic-only
model is the better of the pair by a margin of only $0.07$, and we therefore read MSIS as a tie between our two models rather than as a win.

EXAONE Demand wins $91.9\%$ of its $836$ pairwise comparisons against the other $38$ rows, where
the strongest baseline wins $82.5\%$ of the comparisons among its own peers. The synthetic-only
model wins $88.0\%$ and stays ahead of every baseline in the table, which shows that the synthetic corpus can carry a model on its own.

\subsection{Results by Dataset}
\label{sec:exp_perds}

An aggregate can hide a group of datasets on which a model loses while it still wins on the
geometric mean, and a demand model is exactly the case where that matters, since the hard classes
are a minority of the suite. Table~\ref{tbl:perds} therefore reports every one of the datasets
separately rather than folded into a single number.

The spread across the rows of that table is wider than the spread across its columns, since some datasets are several times harder than the rest for every baseline we ran, and a few
rows of that kind move the aggregate further than any architecture choice of Section~\ref{sec:moe} moves it, as Bitbrains (random), M4 yearly, and M4 daily show.

Neither of our models is last on any of the 22 datasets, each of them is ahead of all 36 baselines
on 7 of them, and the real-world model is ahead of the synthetic-only one on 16 of the 22. The gap
between the two is therefore a lead held across the suite rather than the work of a handful of rows
on which one of them happens to do unusually well, and the same holds of the gap between either of
them and the strongest of the 36 baselines in Table~\ref{tbl:main}.

\begin{table}[t]
\centering
\caption{\textbf{Per-dataset MASE on the 22 evaluation datasets.} Lower is better, the letter in
brackets is the sampling interval, and the five baselines are the strongest of the 36 by their
aggregate over the suite. Real+Synth is the model trained on real-world and
synthetic demand, while Synth is the one trained on the synthetic corpus alone.}
\label{tbl:perds}
\vspace{1mm}
\setlength{\tabcolsep}{3pt}
\adjustbox{max width=\linewidth}{%
\begin{tabular}{l !{\color{black}\vrule} c c !{\color{black}\vrule} c c c c c}
\toprule
\multirow{2.5}{*}{\textbf{Dataset}} & \multicolumn{2}{c|}{\textbf{EXAONE Demand}} & \multicolumn{5}{c}{\textbf{TSFM Baseline}} \\
\cmidrule(lr){2-3} \cmidrule(lr){4-8}
 & \textbf{Real+Synth} & \textbf{Synth} & \textbf{TiRex-1.1}~\citep{auer2025tirex}
& \textbf{Chronos-2}~\citep{ansari2025chronos2} & \textbf{TimesFM-2.5}~\citep{das2024decoder}
& \textbf{Chronos-Bolt (Base)}~\citep{ansari2024chronos} & \textbf{Timer-S1}~\citep{liu2024timer} \\
\midrule
Bitbrains (fast storage) (H) & 1.020 & \second{1.008} & 1.016 & \best{0.995} & 1.120 & 1.082 & 1.126 \\
Bitbrains (random) (H) & \best{5.829} & \second{5.832} & 5.859 & 5.853 & 5.951 & 5.926 & 5.955 \\
BizITObs L2C (H) & 0.441 & 0.445 & 0.487 & 0.439 & 0.520 & \second{0.428} & \best{0.407} \\
Car parts (M) & \best{0.828} & \second{0.847} & 0.916 & 0.900 & 0.942 & 0.903 & 0.893 \\
Electricity (D) & \best{1.375} & \second{1.383} & 1.466 & 1.415 & 1.436 & 1.513 & 1.456 \\
Electricity (H) & 1.010 & 1.009 & 0.912 & \second{0.888} & 0.942 & \best{0.886} & 0.897 \\
Electricity (W) & 1.494 & \second{1.468} & 1.478 & \best{1.456} & 1.491 & 1.517 & 1.614 \\
Hierarchical sales (D) & \best{0.746} & \second{0.761} & 0.768 & 0.768 & 0.786 & 0.765 & 0.776 \\
Hierarchical sales (W) & \best{0.719} & \second{0.722} & 0.747 & 0.745 & 0.748 & 0.764 & 0.755 \\
Hospital (M) & \best{0.759} & 0.765 & 0.785 & 0.807 & \second{0.763} & 0.828 & 0.839 \\
Loop Seattle (D) & \second{0.889} & 0.891 & 0.900 & 0.928 & \best{0.881} & 0.909 & 0.896 \\
Loop Seattle (H) & 0.817 & 0.819 & 0.820 & \second{0.816} & 0.880 & 0.898 & \best{0.808} \\
M4 daily (D) & 3.079 & \best{3.046} & \second{3.063} & 3.164 & 3.202 & 3.070 & 3.336 \\
M4 hourly (H) & 0.763 & 0.805 & \best{0.708} & 0.800 & \second{0.724} & 0.874 & 0.797 \\
M4 monthly (M) & \best{0.909} & \second{0.925} & 0.938 & 0.943 & 0.963 & 0.978 & 1.038 \\
M4 quarterly (Q) & \second{1.149} & 1.150 & \best{1.118} & 1.177 & 1.185 & 1.210 & 1.259 \\
M4 weekly (W) & 2.066 & 2.150 & \best{1.973} & 2.070 & \second{1.980} & 2.154 & 2.442 \\
M4 yearly (Y) & \best{3.165} & \second{3.217} & 3.238 & 3.256 & 3.635 & 3.328 & 3.692 \\
M-dense (D) & 0.645 & \second{0.640} & 0.678 & 0.681 & 0.652 & 0.676 & \best{0.631} \\
M-dense (H) & 0.786 & \second{0.783} & 0.802 & 0.807 & 0.830 & 0.799 & \best{0.776} \\
Restaurant (D) & \best{0.676} & \second{0.677} & 0.704 & 0.713 & 0.696 & 0.730 & 0.717 \\
SZ taxi (H) & \best{0.566} & \second{0.567} & 0.582 & 0.591 & 0.573 & 0.580 & 0.598 \\
\midrule
\rowcolor{LightYellow} $1^{\text{st}}$ Count & 10 & 1 & 3 & 2 & 1 & 1 & 4 \\
\rowcolor{LightYellow} $2^{\text{nd}}$ Count & 2 & 13 & 1 & 2 & 3 & 1 & 0 \\
\bottomrule
\end{tabular}}
\end{table}

\subsection{Forecast Visualization}
\label{sec:exp_examples}

\begin{figure}[t]
\centering
\includegraphics[width=\linewidth]{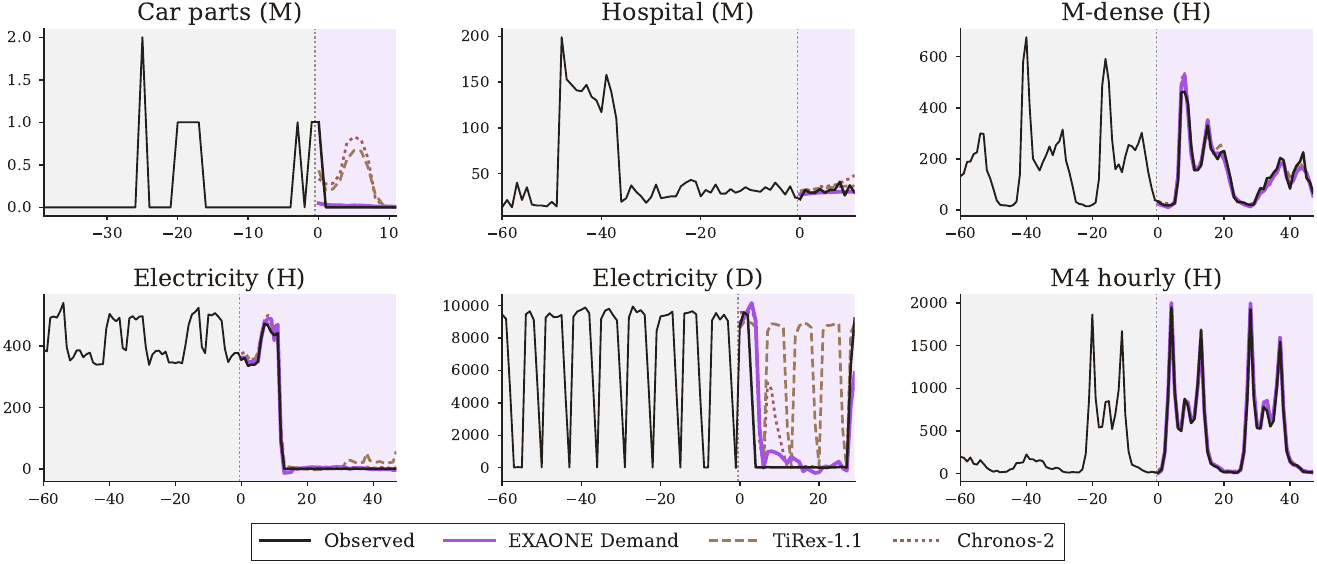}
\vspace{1mm}
\caption{\textbf{Forecasts on six demand series.} Each line is the median forecast of one model,
and EXAONE Demand reaches the lowest MASE of the three on all six series. On Car
parts (M) and Electricity (D), where demand falls to zero, TiRex-1.1 and Chronos-2 keep
forecasting a positive level, while EXAONE Demand follows the zeros.}
\label{fig:examples}
\vspace{1mm}
\end{figure}

Figure~\ref{fig:examples} places EXAONE Demand beside the two strongest baselines of
Table~\ref{tbl:main}, TiRex-1.1 and Chronos-2, on six series of the evaluation suite, and our model
reaches the lowest MASE of the three on every one of them. On the spare-part and hospital series, the two baselines forecast a return of demand that never
arrives, while our model stays at zero on the first and holds the level on the second. On the two
electricity series, demand collapses to zero within the horizon and TiRex-1.1 keeps repeating the cycle
it has read, while our model follows the collapse. On the two hourly series, M-dense and M4 hourly,
our model tracks the height of the sharp peaks most closely.
The panels are chosen to show what the adaptation does when it works, and a selected curve cannot carry the claim that the aggregate carries.

\subsection{Effect of Training Data}
\label{sec:exp_data}

\begin{figure}[t]
\centering
\includegraphics[width=\linewidth]{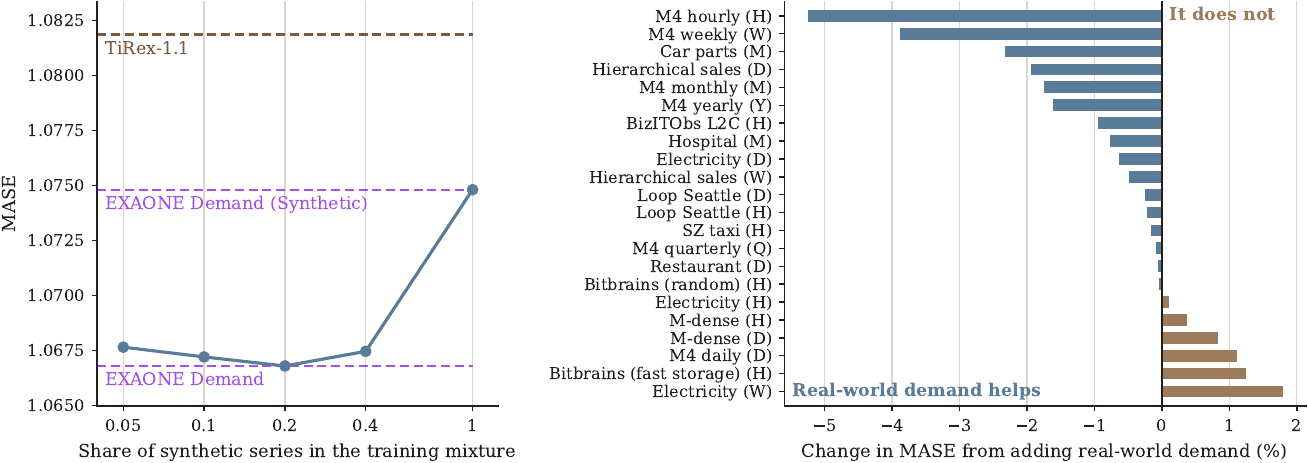}
\vspace{1mm}
\caption{\textbf{What the training data contributes.} [Left] The share of synthetic series in the
mixture, where each point is the median of three seeds except the synthetic-only point at $1.0$,
which is the median of its own runs. [Right] The change in MASE from adding
real-world demand to the synthetic corpus, on the 22 datasets in the evaluation suite.}
\label{fig:data}
\end{figure}

To isolate what real-world demand contributes, we compare runs that share the architecture,
learning rate, target set, and seed, and differ only in whether \textit{real-world series take part
in training}. The left panel of Figure~\ref{fig:data} sweeps the share of
synthetic series in the mixture from $0.05$ to $1.0$, where the last setting is the synthetic corpus
alone. The curve has an interior minimum at the share we adopt, and it rises sharply once no
real-world series is left in the mixture, yet both of our models stay well below TiRex-1.1, the
strongest baseline of Table~\ref{tbl:main}. The right panel places our two models side by
side on each of the 22 datasets, where real-world demand lowers the error on 16 of them and raises
it on 6, and the datasets it helps most are helped several times as much as the ones it costs
anything on are hurt. What the real-world corpus buys is therefore held across the suite rather
than produced by a handful of rows.

\section{Conclusion}
\label{sec:conclusion}

We took a general TSFM as given and asked what a corpus and an adapter built for \textit{demand} are
worth. The corpus is assembled from real-world sources, verified against outside counts, separated
from the evaluation data at the level of \textit{raw values}, and completed by a synthetic generator.
On this corpus, EXAONE Demand trains an MoE adapter whose low-rank branches are mixed by the
demand behaviour of each series, and
both versions, one trained on real-world and synthetic demand and one on synthetic data
alone, outperform every compared TSFM on 22 held-out datasets.

\textbf{Limitation and future works.}
The router reads the context window alone, and the events that drive demand, namely promotions,
holidays, and price changes, reach the model only through the trace they leave in the series, which
makes a covariate-aware router the natural next step. The four classes and the number of routed
branches are fixed rather than learned, and stock-outs are modelled only by the synthetic generator,
which leaves the recovery of latent demand from censored sales open for the next version of the model, together with a router that is supervised by forecast error.

\bibliographystyle{plain}
\bibliography{refs}

\clearpage
\appendix
\section{Hyperparameters}
\label{app:hparams}

This appendix records every setting behind the runs of \S\ref{sec:exp}, so that the model of
\S\ref{sec:moe} can be reproduced from the description alone. Table~\ref{tbl:hparams} lists them in
four groups, namely the frozen backbone, the adapter of Equation~\ref{eq:molora}, the router and its
supervision, and the optimisation. Every value is held fixed across runs, and the data mixture of \S\ref{sec:exp_data} is the one axis that we
sweep across the whole study, at four shares of synthetic series and three seeds for each.

\begin{table}[h]
\centering
\caption{\textbf{Hyperparameters of EXAONE Demand.} The backbone group describes the frozen model
that we adapt, and the remaining three groups describe what we add to it and how it is trained on the demand corpus.}
\label{tbl:hparams}
\vspace{1mm}
\begin{NiceTabular}{l l l}
\toprule
\textbf{Group} & \textbf{Setting} & \textbf{Value} \\
\midrule
\multirow{8}{*}{\makecell[l]{\textbf{Backbone}\\(frozen)}}
 & Blocks                    & 24 \\
 & Model width               & 1{,}024 \\
 & Embedding width           & 2{,}048 \\
 & Attention heads           & 16 \\
 & Context length            & 8{,}192 \\
 & Input and output patch    & 16 \\
 & Output patches per step   & 4 \\
 & Quantile levels           & 21, from $0.01$ to $0.99$ \\
\midrule
\multirow{6}{*}{\makecell[l]{\textbf{Adapter}}}
 & Adapted projections       & $W_q, W_k, W_v, W_o$, and the two feed-forward projections \\
 & Adapted layers            & $24 \times 6 = 144$ \\
 & Branches $E$              & 5, namely 1 shared and 4 routed \\
 & Shared rank $r_0$         & 16 \\
 & Routed rank $r_1 \dots r_4$ & 4 \\
 & Scale $\alpha$, dropout   & 8, $0.05$ \\
\midrule
\multirow{7}{*}{\makecell[l]{\textbf{Router and}\\\textbf{supervision}}}
 & Router width $h$          & 32 \\
 & Router temperature $\tau$ & $1.0$ \\
 & Router learning-rate multiplier & 100 \\
 & Shared weight $w$         & $0.5$ \\
 & Soft cuts $\tau_a, \tau_c$ & $0.35$, $0.60$ \\
 & Sharpening $T$            & $0.5$ \\
 & Auxiliary weight $\lambda$ & $0.1$ \\
\midrule
\multirow{7}{*}{\makecell[l]{\textbf{Optimisation}}}
 & Steps                     & 9{,}000 \\
 & Learning rate             & $1.5 \times 10^{-5}$ \\
 & Weight decay              & $0.1$ \\
 & Warm-up steps             & 300 \\
 & Batch size per device     & 12, accumulated over 4 steps \\
 & Devices and precision     & 8 GPUs with distributed data parallel (DDP), \texttt{bf16} mixed \\
 & Validation interval       & Every 1{,}500 steps \\
\bottomrule
\end{NiceTabular}
\end{table}

The backbone is loaded from its zero-shot checkpoint, and no weight of it receives a gradient at any
point. The router learning-rate multiplier deserves a note of its own,
because the router is the one part of the model that has to move far from its initialisation within
the same budget as the branches. At the base learning rate it barely leaves the uniform mixture,
which produces the collapse described in \S\ref{sec:moe_train} without any other symptom in training.

\clearpage
\section{Related Work}
\label{app:related}

\textbf{Demand forecasting.}
Demand forecasting has a literature of its own, in which intermittent demand is handled by
estimators that forecast the size of a non-zero demand and the interval between two of them
separately~\citep{croston1972forecasting, syntetos2001accuracy}, and a series is placed in one of four
classes by its average inter-demand interval and by the dispersion of its non-zero
values~\citep{syntetos2005categorization}. Classical statistical models fit one estimator per
series~\citep{box1970time}, while forecasting competitions and deep global models fit one model to a
collection of related series~\citep{makridakis2022m5, salinas2020deepar}, and energy load forecasting
has followed the same path~\citep{hong2016probabilistic}. Our router is supervised by the classical
taxonomy, and it carries the separation between classes into a foundation model without giving up
the single set of weights that makes one useful in practice.

\textbf{General time series forecasting models.}
Task-specific forecasting models are trained on one dataset at a time, where patch-based
transformers made long contexts affordable~\citep{zhou2021informer, nie2023patchtst}, and linear,
convolutional, and state space models showed that much of the gain survives without
attention~\citep{zeng2023dlinear, chen2023tsmixer, das2023tide, lee2024pits, bai2018tcn,
luo2024moderntcn, gu2024mamba, wang2024smamba, lee2024sormamba}. Around these models, prior work has studied what the loss should
reward~\citep{lee2026beyond} and how to generate series~\citep{lee2024ant}, each on one dataset at a time.

\textbf{Time series foundation models.}
TSFMs are pretrained on a corpus drawn from many domains and transfer to series that they have never
seen~\citep{garza2023timegpt, ansari2024chronos, woo2024unified, das2024decoder, rasul2023lagllama,
liu2024timer, goswami2024moment, shi2024timemoe, liu2025moiraimoe, ekambaram2024ttm, liu2025sundial,
cohen2024toto, auer2025tirex, ansari2025chronos2, lee2026dataset},
and they are ranked by benchmarks assembled in the same way~\citep{godahewa2021monash,
aksu2024gifteval}. Mixture-of-experts layers~\citep{shazeer2017outrageously} have entered TSFMs as
a way to scale capacity~\citep{shi2024timemoe, liu2025moiraimoe}, whereas we use a mixture of
low-rank experts~\citep{zadouri2024pushing} with a shared expert~\citep{dai2024deepseekmoe} to
specialise one frozen backbone by demand behaviour. Synthetic data is the standard answer to a corpus that cannot be collected at
the required scale~\citep{ansari2024chronos, xie2025cauker}, and specialising a general backbone to
one domain has been carried furthest in finance~\citep{zhu2025fincast, lee2026exaonefinance}, where a
companion effort to ours adapts the same backbone. The present report is its demand-side
counterpart, where the properties that a general corpus under-represents are intermittency, short
histories, censoring, and exogenous events.

\vspace{30pt}
\section{Conversion Schema and Verification}
\label{app:convert}

This appendix records the schema that every source is converted into, the per-series index that is
built alongside it, and the checks that a converted source is required to pass before it is admitted
to the corpus, as referred to in \S\ref{sec:prep}.

All sources are converted to a single schema of \texttt{(series\_id, timestamp, target)}, and we
build a per-series index next to the values themselves, since the checks below compare counts and need every series described first. The
index records three kinds of information for every series:

\begin{itemize}[topsep=1pt,itemsep=0.5pt,parsep=0pt]
\item \textbf{Shape.} Length, observation count, zero ratio, and missing ratio.
\item \textbf{Demand character.} Average inter-demand interval, squared coefficient of
variation, and the resulting Syntetos--Boylan class, all
computed on the converted series rather than on the raw file as it was published.
\item \textbf{Provenance.} Source, domain, licence code, commercial-use flag, and split role.
\end{itemize}

Verification compares the converted counts against numbers established \textit{outside} the
pipeline, such as competition documentation, API totals, or the published number of entities,
rather than against the conversion log. Four quantities are compared for every source, namely the
number of series, the number of observations, the range of timestamps, and the number of distinct
entities where the publisher states one. A source is admitted to the corpus once all four agree
with the published figures, and the per-series index above is what the first two of them are counted
from.

Timestamps are taken from the columns that the source itself carries, and no index is synthesised
from an assumed start date. Where a source renames its columns
between releases, as NHS prescribing data does in turning \texttt{STP\_CODE} into \texttt{ICB\_CODE},
the names are unified before conversion so that every release reaches the same schema.

\clearpage

\section{Domain Breakdown of the Corpus}
\label{app:domains}

Figure~\ref{fig:domains} shows the corpus by series count, and Table~\ref{tbl:domains} gives the
same breakdown numerically while adding the observation count, which tells a different story. The two views disagree because the domains differ in how their
series are shaped, and a domain that is large by one of the two measures can be small by the other.

Energy and transport hold \textit{few but very long} series sampled at high frequency, such as meter
readings and road sensors. Retail and health hold \textit{many but short} ones, since demand there is
recorded per SKU or per practice over a limited window. Energy therefore takes 67.5\% of all
observations from 39.9\% of the series, whereas retail takes 17.2\% of the series and only 1.3\% of the observations, a gap of fifty times on observations against two on series.

\textbf{How a domain is assigned.} Most sources carry one domain, but the six general-purpose
corpora of \S\ref{sec:real_general} hold many datasets under a single name, and a domain assigned to
the corpus would say nothing about the series inside it. The domain is recoverable from the series
identifier, which keeps the name of its source dataset, and we map those 153 dataset names onto the
same domain vocabulary that the rest of the corpus uses, so that \textit{no series is left without a
domain}. The mapping rests on three kinds of evidence: 1) The top-level folders of Time-300B, which
are already domain names such as energy, transport, and sales, 2) the per-dataset domain fields
recorded when subsets of fev-bench were selected, and 3) the published description of each remaining
dataset, read one at a time, which maps \texttt{dominick} and \texttt{rossmann} to retail,
\texttt{electricity} to energy, and \texttt{azure\_vm\_traces} to compute. The mapping changes only
how series are reported in this appendix, and never which of them are used for training the model.

\begin{table}[h]
\centering
\caption{\textbf{Corpus composition by domain.} Shares are over the whole converted corpus, including
the synthetic portion. Series that arrived inside the general-purpose corpora are counted under the
domain of the dataset they came from, not under the corpus that delivered them, so that the shares
reflect the data rather than its packaging.}
\label{tbl:domains}
\vspace{1mm}
\begin{NiceTabular}{l r r r r}
\toprule
\textbf{Domain} & \textbf{Series} & \textbf{Share} & \textbf{Observations} & \textbf{Share} \\
\midrule
Energy                   & 4{,}507{,}842 & 39.87\% & 32{,}646{,}256{,}620 & 67.51\% \\
Retail                   & 1{,}942{,}682 & 17.18\% &   641{,}490{,}811 &  1.33\% \\
Transport                & 1{,}578{,}373 & 13.96\% & 9{,}742{,}740{,}721 & 20.15\% \\
Compute and cloud ops    & 1{,}133{,}431 & 10.02\% & 3{,}686{,}572{,}872 &  7.62\% \\
Web traffic              &     880{,}755 &  7.79\% & 1{,}348{,}375{,}729 &  2.79\% \\
Health                   &     817{,}901 &  7.23\% &    32{,}301{,}783 &  0.07\% \\
Economics                &     372{,}339 &  3.29\% &   100{,}580{,}899 &  0.21\% \\
Synthetic                &      47{,}500 &  0.42\% &    59{,}638{,}451 &  0.12\% \\
Telecom                  &      20{,}000 &  0.18\% &    90{,}925{,}312 &  0.19\% \\
Tourism                  &       3{,}576 &  0.03\% &         460{,}340 &  0.00\% \\
Logistics                &       2{,}065 &  0.02\% &     5{,}790{,}260 &  0.01\% \\
Public services          &           276 &  0.00\% &         303{,}449 &  0.00\% \\
\midrule
Total                    & 11{,}306{,}740 & 100.00\% & 48{,}355{,}437{,}247 & 100.00\% \\
\bottomrule
\end{NiceTabular}
\end{table}

\end{document}